%% file: main.tex
\documentclass[10pt,twocolumn,letterpaper]{article}
\usepackage{iccv}
\usepackage{times}
\usepackage{epsfig}
\usepackage{graphics, graphicx}
\usepackage{amsmath}
\usepackage{amssymb}
\usepackage{xcolor}
\usepackage{booktabs}
\usepackage{amsfonts}
\usepackage{flushend}
\usepackage{algorithm, algorithmic}
\usepackage{url}

\newenvironment{myitemize}{\begin{list}{$\bullet$}
{\setlength{\topsep}{1mm}
\setlength{\itemsep}{0.25mm}
\setlength{\parsep}{0.25mm}
\setlength{\itemindent}{0mm}
\setlength{\partopsep}{0mm}
\setlength{\labelwidth}{15mm}
\setlength{\leftmargin}{4mm}}}{\end{list}}

\newcommand{\lixu}[1]{{{\color{blue} \textit{(Lixu: #1)}}}}

\usepackage[pagebackref=true,breaklinks=true,letterpaper=true,colorlinks,bookmarks=false]{hyperref}

\iccvfinalcopy 

\def\iccvPaperID{2249} 

\ificcvfinal\fi

\begin{document}

\title{Weak Adaptation Learning:\\ Addressing Cross-domain Data Insufficiency with Weak Annotator}

\author{Shichao Xu~\footnotemark[1], Lixu Wang~\footnotemark[1], Yixuan Wang, Qi Zhu\\
Northwestern University, Evanston, USA\\
{\tt\small \{shichaoxu2023, lixuwang2025, yixuanwang2024\}@u.northwestern.edu, qzhu@northwestern.edu}
}

\maketitle
\renewcommand{\thefootnote}{\fnsymbol{footnote}}
\footnotetext[1]{These authors contributed equally to this work.}
\ificcvfinal\thispagestyle{empty}\fi

\begin{abstract}
Data quantity and quality are crucial factors for data-driven learning methods. In some target problem domains, there are not many data samples available, which could significantly hinder the learning process. While data from similar domains may be leveraged to help through domain adaptation, obtaining high-quality labeled data for those source domains themselves could be difficult or costly. 
To address such challenges on data insufficiency for classification problem in a target domain, we propose a \emph{weak adaptation learning (WAL)} approach that leverages unlabeled data from a similar source domain, a low-cost weak annotator that produces labels based on task-specific heuristics, labeling rules, or other methods (albeit with inaccuracy), and a small amount of labeled data in the target domain.  
Our approach first conducts a theoretical analysis on the error bound of the trained classifier with respect to the data quantity and the performance of the weak annotator, and then introduces a multi-stage weak adaptation learning method to learn an accurate classifier by lowering the error bound.
Our experiments demonstrate the effectiveness of our approach in learning an accurate classifier with limited labeled data in the target domain and unlabeled data in the source domain. 
\end{abstract}

\input{section/introduction}
\input{section/related_work}
\input{section/error_bound}

\input{section/method}

\input{section/evaluation}
\input{section/conclusion}

\input{section/acknowledgement}

{\small
\bibliographystyle{ieee_fullname}
\bibliography{shichao_bib}
}

\clearpage
\newpage
\noindent\textbf{\Large{Appendix}}
\appendix
\noindent\input{section/Appendix}

\end{document}

%% file: section/introduction.tex
\section{Introduction}
\label{sec:introduction}

Machine Learning (ML) techniques, especially those based on deep neural networks, have shown great promises in many applications, to a large extent due to their abilities in studying and memorizing the knowledge embedded in high-quality training data~\cite{goodfellow2016deep}.
Having a large number of data samples with accurate labels could enable effective supervised learning methods for improving ML model performance. 
However, it may be difficult to collect many data samples in some problem domains or scenarios,  
such as for the training of
autonomous vehicles during extreme weather (e.g., fog, snow, hail) and natural disasters (e.g., mudflow), or for search and rescue robots during forest fire and earthquake.
One possible solution to such problem of data unavailability is using data from other similar domains to train the target domain model and then fine-tune it with limited target domain data, i.e., through domain adaptation. 
Taking the aforementioned cases as examples, while there may not be much data in hailing weather, we could collect data in days with heavy rain; while it may be difficult to find images during earthquakes for large parts of America, we could collect images in Japan, where earthquakes occur more often in a different environment. 
However, obtaining a large amount of high-quality labeled data in these source domains could still be challenging and costly.  

To address the above data insufficiency challenges across domains, we consider leveraging low-cost weak annotators that can automatically generate large quantity of labeled data based on certain labeling rules/functions,  task-specific heuristics, or other methods (which may be inaccurate to some degree). 
More specifically, our approach considers the following setting for classification problems: There is a small amount of data samples with accurate labels collected for the target domain, which is called \emph{target domain data} or \emph{target data} in this paper for simplicity. There is also a large amount of unlabeled data that can be acquired from a similar but different source domain (i.e., there exists domain discrepancy), which is called \emph{source (domain) data} in this paper. Finally, there is a weak annotator that can produce weak (possibly inaccurate) labels on data samples. Our objective is to learn an accurate classifier for the target domain based on the labeled target data, the initially-unlabeled source data, and the weak annotator. 

The problem we are considering here is related but different from Semi-Supervised Learning (SSL)~\cite{rasmus2015semi, dong2018tri, li2019dividemix} and Unsupervised Domain Adaptation (UDA)~\cite{li2020model, dong2020can, wilson2020survey, dong2019semantic}. In the setting of SSL, the available training data consists of two parts -- one has accurate labels while the other is unlabeled, and the two parts are drawn from the same distribution in terms of training features. This is different from our problem, where there exists domain discrepancy across the source and target domains. 
The objective of UDA is to adapt a model to perform well in the target domain based on labeled data in the source domain and unlabeled data in the target domain. This is again very different from our problem, where the source domain data is initially unlabeled and assigned with inaccurate labels by a weak annotator, while the target domain data has labels but its quantity is small. 
Another related field is Positive-unlabeled Learning (PuL)~\cite{kiryo2017positive, chen2020self}, an approach for sample selection. The training data of PuL also consists of two parts -- positive and negative data, and the task is to learn a binary classifier to filter out samples that are similar to the positive data from a large amount of negative data. However, the current PuL approaches usually conduct experiments in a single data set rather than multiple domains with feature discrepancy.

To solve our target problem, we first develop a theoretical analysis on the error bound of a trained classifier with respect to the data quantity and the weak annotator performance. We then propose a Weak Adaptation Learning (WAL) method 
to learn an accurate classifier by lowering the error bound. The main idea of WAL is to obtain a cross-domain representation for both source domain and target domain data, and then use the labeled data to estimate the classification error/distance between the weak annotator and the ideally optimal classifier in the target domain. Next, all the data is re-labeled based on such estimation of weak annotator classification error. Finally, the newly-relabeled data is used to learn a better classifier in the target domain. 

Our work makes the following contributions:
\begin{myitemize}
    \item We address the challenge of data insufficiency in domain adaptation with a novel weak adaptation learning approach that leverages unlabeled source domain data, limited number of labeled target domain data, and a weak annotator.
    \item Our approach includes a theoretical analysis on the error bound of the trained classifier and a multi-stage WAL method that improves the classifier accuracy by lowering such error bound.
    \item We compare our approach with various baselines in experiments with domain discrepancy setting on several digit datasets and the VisDA-C dataset, and study the cases without domain discrepancy on the CIFAR-10 dataset. We also conduct ablation studies on the impact from the weak annotator accuracy and the quantity of labeled data samples to further validate our ideas.
\end{myitemize}

%% file: section/related_work.tex
\section{Related Work}
\label{sec:related_work}
We introduce related works in the topics about weakly- and semi-supervised learning, and the importance of sample quantity here. You can also find more related works about domain adaptation in the supplementary materials.
\subsection{Weakly- and Semi-Supervised Learning}%

Weakly Supervised Learning is a large concept that may have multiple problem settings~\cite{zhou2018brief}. The problem we consider in this paper is related to the incomplete supervision setting that is often addressed by Semi-Supervised Learning (SSL) approaches.
Standard SSL solves the problem of training a model with a few labeled data and a large amount of unlabeled data. Some of the widely-applied methods~\cite{rasmus2015semi, dong2018tri, perez2018weakly, belharbi2020deep} assign pseudo labels to unlabeled samples and then perform supervised learning. And there are works that address the noises in the labels of those samples~\cite{natarajan2013learning, ghosh2015making, liu2015classification}. 
Our target problem is related to SSL with inaccurate supervision, but is different since we consider the feature discrepancy between the (unlabeled) source data and the (labeled) target data -- a case that occurs often in practice but has not been sufficiently addressed.  

Positive-unlabeled Learning (PuL) is usually regarded as a sub-problem of SSL. Its goal is to learn a binary classifier to distinguish positive and negative samples from a large amount of unlabeled data and a few positive samples. Several works~\cite{kiryo2017positive, chen2020self} can achieve great performance on selecting samples that are similar to the positive data, and there are also works using samples selected by PuL to perform other tasks~\cite{xu2019positive, loghmani2020positive}.

\subsection{Importance of Sample Quantity}
The training of machine learning models, especially deep neural networks, often requires a large amount of data samples. However, in many practical scenarios, there is not sufficient training data to feed the learning process, degrading the model performance sharply~\cite{snell2017prototypical, johnson2019survey, xu2021learningbased}. Many approaches have been proposed to make up for the lack of training samples, e.g., data re-sampling~\cite{van2007experimental}, data augmentation~\cite{pouyanfar2018dynamic}, metric learning and meta learning~\cite{cao2019theoretical, chen2019closer, triantafillou2019meta, wang2021addressing}. And there are works~\cite{neyshabur2017exploring, amit2018meta, cao2019theoretical, wang2020generalizing} conducting theoretical analysis on the relation between training data quantity and model performance. These analyses are usually in the form of bounding the prediction error of the models and provide valuable information on how the sample quantity of training data affects the model performance. In our work, we also perform a theoretical analysis on the error bound of the trained model, with respect to not only the data quantity but also the performance of the weak annotator.


%% file: section/error_bound.tex
\section{Theoretical Analysis}
\subsection{Problem Definition and Formulation}

We consider the task of classification, where the goal is to predict labels for samples in the target domain. Two types of supporting data can be accessed for training the model -- source domain data and target domain data. The source domain data samples are initially unlabeled and come from a joint probability distribution $\mathbb{Q}^s$.  
They can be labeled by a weak annotator $\mathbf{h}^w$ (which may be inaccurate) and denoted as $D_s = \{(\mathbf{x}_s,y_s)_i\}_{i=1}^{N_s}$, where $N_s$ is the number of source data samples. The target domain data $D_t = \{(\mathbf{x}_t, y_t)_i\}_{i=1}^{N_t}$ consists of $N_t$ samples collected from the target distribution $\mathbb{Q}^t$. Note that $\mathbb{Q}^t$ may be different from $\mathbb{Q}^S$. And we use $\mathbb{Q}^s_X$, $\mathbb{Q}^s_Y$ and $\mathbb{Q}^t_X$, $\mathbb{Q}^t_Y$ to represent the marginal distributions of the source and target domains, respectively.  Moreover, as stated before, we consider the case where there is only a small amount of target domain data, i.e., $N_t \ll N_s$.




Our goal is to learn an accurate classifier for the target domain. The classifier is initialized from a parameter distribution $\mathcal{H}$, which denotes the hypothesis parameter space of all possible classifiers.

In the following analysis, we will define the classification risk of a classifier and then derive its bound. According to the PAC-Bayesian framework~\cite{mcallester1999some, germain2016pac}, the expected classification risk of a classifier drawn from a distribution $\mathcal{Q}$ that depends on the training data can be strictly bounded. Let $\mathbf{h}_{\Theta}$ denote a learned classifier from the training data, and its parameter $\Theta$ is drawn from $\mathcal{Q}$. We consider that the prior parameter distribution $\mathcal{H}$ over the hypothesis is independent of the training data. And given a $\delta$ with the probability $\ge 1-\delta$ over the training data set of size $m$, the expected error of $\mathbf{h}_{\Theta}$ can be bounded as follows~\cite{mcallester2003simplified}:
\begin{equation}
\setlength{\abovedisplayskip}{3pt}
\setlength{\belowdisplayskip}{3pt}
\begin{aligned}
    L(\mathbf{h}_{\Theta}) &\leq \widehat{L}(\mathbf{h}_{\Theta})+\sqrt{\widehat{L}(\mathbf{h}_{\Theta}) \cdot \Omega} + \Omega \\
    \Omega &= \frac{2\left(K\!L(\mathcal{Q} \| \mathcal{H})+\ln \frac{m}{\delta}\right)}{m-1}
\end{aligned}
\label{pac_theorem}
\end{equation}
Here $L(\mathbf{h}_{\Theta})$ is the expected error of $\mathbf{h}$ over parameter $\Theta$, and $\widehat{L}(\mathbf{h}_{\Theta})$ is the empirical error computed from the training set ($\widehat{L}(\mathbf{h}_{\Theta}) = \frac{1}{m} \sum_{i=1}^m \mathcal{L}(\mathbf{x}_i, y_i)$, where $\mathcal{L}$ denotes the loss of a single training sample). In Eq.~(\ref{pac_theorem}), $K\!L(\mathcal{Q} \| \mathcal{H})$ represents the Kullback-Leibler (KL) divergence between parameter distribution $\mathcal{Q}$ and $\mathcal{H}$. For any two distributions $p, \, q$, the specific form of their KL divergence is $K\!L(p \| q) = -\mathbb{E}[p \cdot \ln{\frac{q}{p}}]$. In most cases of mini-batch training, the training loss $\widehat{L}(\mathbf{h}_{\Theta})$ is much smaller than $\Omega$, and thus we can get a further bound as follows~\cite{neyshabur2017exploring}:
\begin{equation}
    \setlength{\abovedisplayskip}{3pt}
    \setlength{\belowdisplayskip}{3pt}
    L(\mathbf{h}_{\Theta}) \leq \widehat{L}(\mathbf{h}_{\Theta})+4\sqrt{\frac{\left(K\!L(\mathcal{Q} \| \mathcal{H})+\ln \frac{2m}{\delta}\right)}{m}}
    \label{pac_theorem_further}
\end{equation}

Then if we denote the model parameters of $\mathbf{h}$ before the training that are drawn from $\mathcal{H}$ as $\Theta^{\mathbf{p}}$, the KL divergence can be written as 
$K\!L(\mathcal{Q} \| \mathcal{H}) = -\mathbb{E} [ \Theta \cdot (\ln{\Theta^{\mathbf{p}}} - \ln{\Theta}) ]$.
As aforementioned, $\mathbf{h}_{\Theta}$ is trained with the training data set from $\mathbf{h}_{\Theta^{\mathbf{p}}}$, and we consider that the training is optimized by gradient-based method. Thus, we can formulate that $\Theta = \Theta^{\mathbf{p}} + \nabla (\widehat{L}(\mathbf{h}_{\Theta^{\mathbf{p}}}))$. Here we omit the learning rate to simplify the formula.

The PAC-Bayesian error bound is valid for any parameter distribution $\mathcal{H}$ that is independent of the training data, and any method of optimizing $\Theta^{\mathbf{p}}$ dependent on the training set~\cite{neyshabur2017exploring}. Therefore, in order to simplify the problem, we instantiate the bound as setting $\mathcal{H}$ to conform to a Gaussian distribution with zero mean ($\mu_{\mathcal{H}} = 0$) and $\text{Var}_{\mathcal{H}} = \sigma^2_{\mathcal{H}}$ variance. This simplification is the same as previous PAC-Bayesian works~\cite{neyshabur2017exploring, neyshabur2018pac}. We further assume that the parameter change of the overall model during training can also be regarded as conforming to an empirical Gaussian distribution. This Gaussian distribution is independent of model parameters if we regard the parameter updates induced by gradient back-propagation as accumulated random perturbations, i.e., each training sample corresponds to a small perturbation~\cite{neyshabur2018pac}. And we denote the mean and the variance of a single training sample as follows:
\begin{equation}
\setlength{\abovedisplayskip}{3pt}
\setlength{\belowdisplayskip}{3pt}
\begin{aligned}
    \mu &\triangleq \mathbb{E} \left[\nabla_{\Theta^{\mathbf{p}}} \mathcal{L}(\mathbf{x}, y)\right] \\
    \sigma^2 &\triangleq \mathbb{E} \left[(\nabla_{\Theta^{\mathbf{p}}} \mathcal{L}(\mathbf{x}, y) - \mu)(\nabla_{\Theta^{\mathbf{p}}} \mathcal{L}(\mathbf{x}, y) - \mu)^T\right]
\end{aligned}
\label{distribution of learned parameter_main paper}
\end{equation}
Then, the specific formula of KL divergence to any two Gaussian distributions $p \sim \mathcal{N}(\mu_1, \sigma_1^2)$, $q \sim \mathcal{N}(\mu_2, \sigma_2^2)$ is written as follows:
\begin{equation}
    \setlength{\abovedisplayskip}{3pt}
    \setlength{\belowdisplayskip}{3pt}
    K\!L(p, q) = \ln{\frac{\sigma_2}{\sigma_1}} + \frac{\sigma_1^2 + (\mu_1 - \mu_2)^2}{2\sigma_2^2} - \frac{1}{2}
\end{equation}

\noindent \emph{\textbf{Theorem 1.}} \textit{For a classifier parameter distribution $\mathcal{H} \sim \mathcal{N}(0, \sigma_{\mathcal{H}}^2)$ that is independent of the training data with size $m$, and a posterior parameter distribution $\mathcal{Q}$ learned from the training data set, if we assume $\mathcal{Q} \sim \mathcal{N}(\mu_{\mathcal{Q}}, \sigma_{\mathcal{Q}}^2)$ and consider $\Theta^{\mathbf{p}}$, $\Theta$ as drawn from $\mathcal{H}$, $\mathcal{Q}$ respectively ($\Theta = \Theta^{\mathbf{p}} + \nabla (\widehat{L}(\mathbf{h}_{\Theta^{\mathbf{p}}}))$), the KL divergence of $\mathcal{Q}$ and $\mathcal{H}$ is bounded with symbols defined in Eq.~(\ref{distribution of learned parameter_main paper}) as follows:}
\begin{equation}
    K\!L(\mathcal{Q}\, \|\, \mathcal{H}) \le \frac{\frac{\sigma^2}{m} + \mu^2}{2\sigma_{\mathcal{H}}^2}
\end{equation}
The detailed proof of Theorem 1 is presented in our Supplementary Materials. With the above risk definition, the risk of $\mathbf{h}$ with respect to the target data distribution $\mathbb{Q}^t$ is
\begin{equation}
\centering
R^t(\mathbf{h}) = \mathbb{E}_{(\mathbf{x},y) \sim \mathbb{Q}^t} \mathcal{L}(\mathbf{h}(\mathbf{x}), y) = L(\mathbf{h}_{\Theta})_{\sim \mathbb{Q}^t}
\end{equation}
Besides, we define the Classification Distance of two classifiers $\mathbf{h}_1$ and $\mathbf{h}_2$ under the same domain distribution $\mathbb{P}$ as 
\begin{equation}
\mathcal{CD}_{\sim \mathbb{P}}(\mathbf{h}_1, \mathbf{h}_2) = \mathbb{E}_{\mathbf{x} \sim \mathbb{P}} \mathcal{L}(\mathbf{h}_1(\mathbf{x}), \mathbf{h}_2(\mathbf{x}))
\end{equation}
Moreover, the Discrepancy Distance of two domains is defined as in~\cite{mansour2009domain}:  $\forall \mathbf{h_1}, \mathbf{h_2}$, the discrepancy distance between the distributions of two domains $\mathbb{P}, \mathbb{Q}$ is
\begin{equation}
    \setlength{\abovedisplayskip}{3pt}
    \setlength{\belowdisplayskip}{3pt}
    \mathcal{DD} (\mathbb{P}, \mathbb{Q}) = \sup_{\mathbf{h_1}, \mathbf{h_2} \in \mathbb{H}} |\mathcal{CD}_{\sim \mathbb{P}}(\mathbf{h}_1, \mathbf{h}_2) - \mathcal{CD}_{\sim \mathbb{Q}}(\mathbf{h}_1, \mathbf{h}_2)|
    \label{domain discrepancy}
\end{equation}

For further analysis, we also define two operators in a parameter distribution $\mathcal{H}$:
\begin{myitemize}
    \item $\oplus$: 
    $\forall \mathbf{h_1}, \mathbf{h_2} \in \mathcal{H}$, and $\forall \mathbf{x} \in \mathbb{P}$, a new classifier $\mathbf{h_3} = \mathbf{h_1} \oplus \mathbf{h_2}$ can be acquired by conducting operator $\oplus$ on $\mathbf{h_1}$ and $\mathbf{h_2}$, and $\mathbf{h_3}(\mathbf{x}) = \mathbf{h_1}(\mathbf{x}) + \mathbf{h_2}(\mathbf{x})$.
    
    \item $\ominus$: 
    $\forall \mathbf{h_1}, \mathbf{h_2} \in \mathcal{H}$, and $\forall \mathbf{x} \in \mathbb{P}$, a new classifier $\mathbf{h_3} = \mathbf{h_1} \ominus \mathbf{h_2}$ can be acquired by conducting operator $\ominus$ on $\mathbf{h_1}$ and $\mathbf{h_2}$, and $\mathbf{h_3}(\mathbf{x}) = \mathbf{h_1}(\mathbf{x}) - \mathbf{h_2}(\mathbf{x})$.
\end{myitemize}

\subsection{Error Bound Analysis}

Let $\mathbf{h}^{o_s}$ and $\mathbf{h}^{o_t}$ denote the ideal classifiers that perform optimally on the source data and target data, respectively:
\begin{equation}
\begin{aligned}
    \mathbf{h}^{o_s} &= {\arg \min}_{\mathbf{h} \in \mathcal{Q}}{R^s(\mathbf{h})} ,\ 
    \mathbf{h}^{o_t} = {\arg \min}_{\mathbf{h} \in \mathcal{Q}}{R^t(\mathbf{h})}
\end{aligned}
\end{equation}

In our approach, we design a classifier that learns the discrepancy between the weak annotator and the ground truth (details will be introduced in Section~\ref{sec:WAL}), and we denote it as $\mathbf{d}$ drawn from $\mathcal{Q}$. Thus, we can get a model that is the product of conducting the aforementioned $\oplus$ operator on $\mathbf{h}$ and $\mathbf{d}$, i.e., $\mathbf{h} \oplus \mathbf{d}$. Here $\mathbf{h}$ is designed for approximating the weak labels. And for the risk of $\mathbf{h} \oplus \mathbf{d}$, we can obtain the following relation:

\noindent \emph{\textbf{Theorem 2.}} \textit{For all L1 (Mean Absolute Error~\cite{kassam1978quantization}), L2 (Mean Squared Error~\cite{gunst1977biased}) and their non-negative combination loss functions (Huber Loss~\cite{yi2017semismooth}, Quantile Loss~\cite{koenker2004quantile}, etc.), the classification risk of aforementioned $\mathbf{h} \oplus \mathbf{d}$ can be formulated as follows:}
\begin{equation}
\begin{aligned}
    R^t(\mathbf{h} \oplus \mathbf{d})
    &= \mathbb{E}_{\mathbb{Q}^t_X}\mathcal{L}(\mathbf{h} \oplus \mathbf{d}, \mathbf{h}^{w} \oplus \mathbf{h}^{o_t} \ominus \mathbf{h}^w)\\
    &\leq \mathbb{E}_{\mathbb{Q}^t_X}\mathcal{L}(\mathbf{h}, \mathbf{h}^{w}) + \mathbb{E}_{\mathbb{Q}^t_X}\mathcal{L}(\mathbf{d}, \mathbf{h}^{o_t} \ominus \mathbf{h}^w)\\
\end{aligned}
\label{basic term of overall bound}
\end{equation}
Please refer to our Supplementary Materials for detailed proof of Theorem 2. Then if we consider that the training loss $\widehat{L}(\mathbf{h})$ (which equals the average loss of all training samples) is hardly influenced by the sample quantity, and it is the same for the discrepancy between two domains~\cite{luo2020progressive}, we can split the error bound of $\mathbf{h} \oplus \mathbf{d}$ into two parts, where one part, denoted as $\Delta$, is not influenced by the sample quantity and the other is related to the sample quantity. According to Eq.~(\ref{pac_theorem_further}), these two parts can be written as follows (the detailed derivation of inequalities starting from Eq.~(\ref{basic term of overall bound}) can be found in the Supplementary Materials):
\begin{equation}
\begin{aligned}
    R^t(\mathbf{h} \oplus \mathbf{d})
    &= \mathbb{E}_{\mathbb{Q}^t_X}\mathcal{L}(\mathbf{h} \oplus \mathbf{d}, \mathbf{h}^{w} \oplus \mathbf{h}^{o_t} \ominus \mathbf{h}^w)\\
    &\leq \Delta + 4\sqrt{\frac{K\!L_{\mathbf{d}}}{N_t}} + 4\sqrt{\frac{K\!L_{\mathbf{h}}}{N_s}} \\
    &\quad+ 12\sqrt{\frac{\ln{\frac{2N_t}{\delta}}}{N_t}} + 8\sqrt{\frac{\ln{\frac{2N_s}{\delta}}}{N_s}} \\ 
    \text{where}\quad \Delta &= 2\widehat{L}_t(\mathbf{h}^w) + \widehat{L}_t(\mathbf{d}) + \widehat{L}_s(\mathbf{h}) \\
    &\quad+ \widehat{L}_s(\mathbf{h}^w) + \mathcal{DD}(\mathbb{Q}^t_X, \mathbb{Q}^s_X)
\end{aligned}
\label{basic term of overall bound in paper}
\end{equation}
Here $K\!L_{\mathbf{d}}$ and $K\!L_{\mathbf{h}}$ denote KL divergences between trained $\mathbf{d}$, $\mathbf{h}$ and $\mathcal{H}$ respectively. According to Theorem 1, this KL divergence term is influenced by the training, especially impacted by the sample quantity. We will discuss the insights obtained from this error bound in the next section, and then introduce our weak adaptation learning process that is inspired by those insights.

%% file: section/method.tex
\section{Weak Adaptation Learning Method}
\label{sec:WAL}
\subsection{Observation from Error Analysis}
Based on the error bound derived in Eq.~(\ref{basic term of overall bound in paper}), we can put efforts into the following ideas in our approach to improve the classifier performance in the target domain:
\begin{myitemize}
    \item \textbf{Performance of annotator} ($2\widehat{L}_t(\mathbf{h}^w) + \widehat{L}_s(\mathbf{h}^w)$): The supervision provided by the weak annotator can guide the model to better target the given task. Ideally, we want $\mathbf{h}^w$ to produce more accurate labels for both source and target data, reducing $2\widehat{L}_t(\mathbf{h}^w)$ and $\widehat{L}_s(\mathbf{h}^w)$ simultaneously. Practically though, we may just be able to make the annotator perform better on the source domain and cannot do much with the target domain. 
    \item \textbf{Discrepancy between domains} ($\mathcal{DD}(\mathbb{Q}^t_X, \mathbb{Q}^s_X)$): Designing loss to quantify the discrepancy between the source and target domains is well studied in Domain Adaptation. In our approach, we propose a novel inter-domain loss (called Classified-MMD) to minimize $\mathcal{DD}(\mathbb{Q}^t_X, \mathbb{Q}^s_X)$, as introduced later.
    \item \textbf{Quantity of source and target samples} ($N_s, N_t$): First, the learning of $\mathbf{d}$ needs the supervision of the ground truth, and thus we can only use the labeled target data to train $\mathbf{d}$. Then, in our method, $\mathbf{h}$ is designed to approximate the weak annotator, and therefore it may see enough that we just use the source data to train $\mathbf{h}$. However, to further reduce $K\!L_{\mathbf{h}}$ according to Theorem 1, we also use target samples to train $\mathbf{h}$, which increases the sample size of training data.
    Moreover, since the sample quantity of source data is much larger than that of target data (i.e., $N_t \ll N_s$), $\sqrt{K\!L_{\mathbf{d}} / N_t}$ in Eq.~(\ref{basic term of overall bound in paper}) dominates over $\sqrt{K\!L_{\mathbf{h}} / N_s}$, and in the case of $\delta \leq 2 / e$, $12\sqrt{\ln{\frac{2N_t}{\delta}} / N_t}$ also strictly dominates over $8\sqrt{\ln{\frac{2N_s}{\delta}} / N_s}$. As the result, 
    the terms influenced by a few target samples dominates the overall error risk. Therefore, directly applying $\mathbf{h} \oplus \mathbf{d}$ to the target domain will still be impacted by the insufficient samples. However, note that $\mathbf{h} \oplus \mathbf{d}$ can produce more accurate labels for the source data than the weak annotator. Therefore, we add a final step in our learning process that utilizes re-labeled source data and conducts supervised learning with such augmentation.
\end{myitemize}

\subsection{Learning Process}
\begin{figure*}[htb]
\centering
\includegraphics[width=18cm]{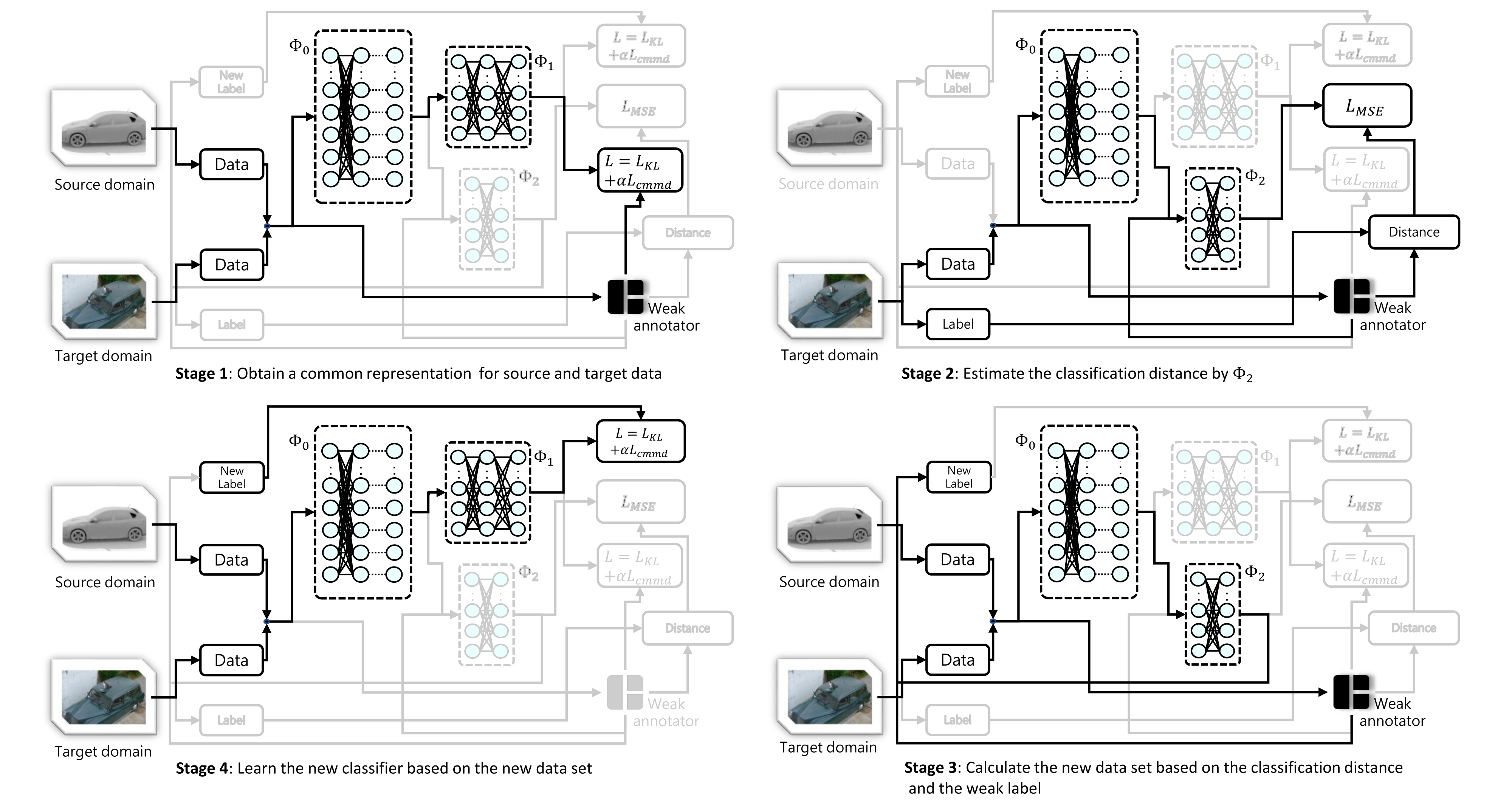}
\caption{Overview of the Weak Adaptation Learning (WAL) process. The designed network architecture is divided into three components $\Phi_0, \Phi_1, \Phi_2$ and the algorithm has four stages. First, we use a combined loss function to learn a cross-domain representation in $\Phi_0$ for both source and target data samples. Then, in Stage 2, $\Phi_2$ estimates the classification distance between the weak annotator and the ideally optimal one in the target domain. A new re-labeled dataset is calculated in Stage 3, and then used in Stage 4 to learn the desired classifier.}
\vspace{-0.2cm}
\label{overview}
\end{figure*}

In this section, we present the detailed process of our 
weak adaptation learning (WAL) method, which is designed based on the observations from the above error bound analysis.  
The overview of our WAL process is shown in Figure~\ref{overview}. The designed network consists of three parts -- ($\Phi_0, \Phi_1, \Phi_2$). $\Phi_0$ can be seen as a shared feature network for both source and target data, using typical classification networks such as VGG, ResNet, etc. $\Phi_1$ consists of three fully-connected layers that follow the output of $\Phi_0$. And we denote the combination of $\Phi_0$ and $\Phi_1$ as $F_1$. $\Phi_2$ consists of two fully-connected layers that follow the output of $\Phi_0$. The combination of $\Phi_0$ and $\Phi_2$ is denoted as $F_2$. The detailed network architecture is shown in the Supplementary Materials. The workflow of our method is shown in Algorithm \ref{algo1}.

\begin{algorithm}[htbp]
\small
\caption{The workflow of Weak Adaptation Learning.}
\label{algo1}
\begin{algorithmic}[1]
\STATE Initialize parameters of network components $\Phi_0, \Phi_1, \Phi_2$.
\STATE Obtain dataset $D$ from the source and target data with the help of weak annotator $\mathbf{h}^w$.
\STATE Train $F_1 = \Phi_1 \circ \Phi_0$ using $D$, with loss function following equation $\mathcal{L} = \mathcal{L}_{K\!L} + \alpha \mathcal{L}_{cmmd}$.
\STATE Fix the parameters of $\Phi_1$ and use $F_2 = \Phi_2 \circ \Phi_0$ to fit the distance of the optimal classifier for target data $\mathbf{h}^{o_t}$ and the weak annotator $\mathbf{h}^w$ with the target data.
\STATE Calculate a new dataset using both source and target data. The new labels are calculated by $y^{new} = \mathbf{h}^w(\mathbf{x}) + \Phi_2(\mathbf{h}^w(\mathbf{x}),$ $\Phi_0(\mathbf{x}))$.
\STATE Initialize parameters of $\Phi_0, \Phi_1, \Phi_2$.
\STATE Fix $\Phi_2$ and train $F_1$ using the new dataset. The loss function follows $\mathcal{L} = \mathcal{L}_{K\!L} + \alpha \mathcal{L}_{cmmd}$.
\STATE Output classifier $F_1$.
\end{algorithmic}
\vspace{-0.0cm}
\end{algorithm}

\textbf{\emph{Stage 1}}: The first goal we step on is to obtain a common representation for both the source and target data, which helps us encode the inputs while mitigating the domain discrepancy in the feature representation. We gather all the unlabeled source data and the target data without their labels and use weak annotator $\mathbf{h}^w$ to assign a label for each data sample $\mathbf{x}_{i}$ and $y^{w}_{i} = \mathbf{h}^w(\mathbf{x}_{i})$. We denote the dataset obtained in this way as $D = \{(\mathbf{x}, y^{w})_i\}_{i=1}^{N_s + N_t}$. Then we fix $\Phi_2$ and only consider the left part of the network, which is $F_1 = \Phi_1 \circ \Phi_0$. It is normally trained by supervised learning using the dataset $D$ for ${ep}_1$ training epochs, and uses the following loss function: 
\begin{equation}
\begin{aligned}
    \mathcal{L} & = \mathcal{L}_{K\!L} + \alpha \mathcal{L}_{cmmd}
\end{aligned}
\end{equation}

In this loss function, there are two loss terms and the hyper-parameter $\alpha$ is a scaling factor to balance the scale of two loss functions (we set it as $0.0001$ in our experiments). The first term $\mathcal{L}_{K\!L}$ is the Kullback-Leibler (KL) divergence loss, stated as follows:
\begin{equation}
\begin{aligned}
    \mathcal{L}_{K\!L} &= K\!L{(y_{pre}^1 \Arrowvert y^{w})} \\
    &= K\!L{(\Phi_1 \circ \Phi_0 (\mathbf{x}) \Arrowvert \mathbf{h}^w(\mathbf{x}))}
\end{aligned}
\end{equation}
where $y_{pre}^1$ is the output prediction value of $F_1$ and $y^{w}$ is the corresponding weak label produced by the weak annotator $\mathbf{h}^w$.
The second term $\mathcal{L}_{cmmd}$ aims to mitigate the domain discrepancy of the source and target domain at the feature representation level in the neural networks. Based on the basic MMD loss introduced by~\cite{tzeng2014deep}, we further change it into the version with data labels. We call this loss function as Classified-MMD loss (corresponding to the subscript $cmmd$), which is defined as:
\begin{equation}
\setlength{\abovedisplayskip}{3pt}
\setlength{\belowdisplayskip}{3pt}
\begin{aligned}
    \mathcal{L}_{cmmd} = \frac{1}{M} \cdot \sum_{i=1}^{M} \Arrowvert & \frac{1}{|D_X^{(S,i)}|}\sum_{\mathbf{x}_s \in D_X^{(S,i)}}{F_1(\mathbf{x}_s)} \\
    & - \frac{1}{|D_X^{(T,i)}|}\sum_{\mathbf{x}_t \in D_X^{(T,i)}}{F_1(\mathbf{x}_t)}\Arrowvert 
\end{aligned}
\end{equation} 
where $M$ is the number of classes, $D_X$ is the data from the produced dataset $D$ without labels, and $D_X^{(S,i)}$ is the source data selected from $D_X$ with $\arg \max (y^{w}) = i$. Then, we utilize target data with its accurate labels to continue to train the network component $F_1$ under the loss function $\mathcal{L}_{K\!L}$ for ${ep}_2$ training epochs, which helps further fine-tune the feature we learned through accurate labels of the target data.

\textbf{\emph{Stage 2}}: After finishing training in Stage 1, the next step is to estimate the distance of the optimal classifier for target data $\mathbf{h}^{o_t}$ and the weak annotator $\mathbf{h}^w$. We estimate this distance through available target data with accurate labels. We adopt the parameters trained from Stage 1 and train network component $F_2 = \Phi_2 \circ \Phi_0$ using the target data $D_t$. For an input data sample $\mathbf{x}$, it is brought into both $\Phi_0$ and the weak annotator as their input. And then $\Phi_2$ takes the output feature of $\Phi_0(\mathbf{x})$ and $\mathbf{h}^w(\mathbf{x})$ as input feature (these two features are concatenated as the input feature of $\Phi_2$). For data sample $(\mathbf{x}_{t}, y_{t}) \in $ target dataset $D_t$, the learning of $F_2$ uses the following classifier discrepancy loss function:
\begin{equation}
\begin{aligned}
    \mathcal{L}_{M\!S\!E} & = {\parallel \Phi_2(\mathbf{h}^w(\mathbf{x}_t), \Phi_0(\mathbf{x}_t)) -  (y_{t} - \mathbf{h}^w(\mathbf{x}_{t}))\parallel}^2
\end{aligned}
\end{equation}
 The network is trained for ${ep}_3$ training epochs.

\textbf{\emph{Stage 3}}: The third step is to generate a new dataset $D_{new}$ through the obtained network $F_2$ above. Specifically, we collect data $\mathbf{x}$ from both source data and target data, and we re-label these data based on the weak annotator and $F_2$ obtained from the previous steps:
\begin{equation}
\begin{aligned}
    D_{new} = \{(&\mathbf{x}, y^{new}) | \mathbf{x} \in D_X, \\
    &y^{new} = \mathbf{h}^w(\mathbf{x}) + \Phi_2(\mathbf{h}^w(\mathbf{x}), \Phi_0(\mathbf{x}))\} 
\end{aligned}
\setlength{\belowdisplayskip}{3pt}
\end{equation}


\textbf{\emph{Stage 4}}: In the last step, we focus on $F_1 = \Phi_1 \circ \Phi_0$ again. We fix the parameters of network component $\Phi_2$ and train $F_1$ using the new dataset $D_{new}$ obtained in Stage 3. To avoid introducing feature bias from the previous steps, we clean all previous network weights and re-initialize the whole network before training. The training lasts for ${ep}_4$ epochs, and the loss function for this step is $\mathcal{L} = \mathcal{L}_{K\!L} + \alpha \mathcal{L}_{cmmd}$, which is the same as the function in Stage 1. Finally, we get the final model $F_1$ as the desired classifier. 

To sum up, in Stage 1, we learn the model $\mathbf{h}$ with the help of the weak annotator to decrease the empirical loss $\widehat{L}_s(\mathbf{h})$, and the CMMD loss will reduce the term $\mathcal{DD}(\mathbb{Q}^t_X, \mathbb{Q}^s_X)$. Stage 2 uses a new classifier $\mathbf{d}$ to learn the classification distance corresponding to the term $\widehat{L}_t(\mathbf{d})$. The Stage 3 uses the annotator and the learned $\mathbf{d}$ to give more accurate labels than those given solely by the annotator. Then in Stage 4, the model is trained by the relabeled data, making both $\widehat{L}_s(\mathbf{h})$ and $\mathcal{DD}(\mathbb{Q}^t_X, \mathbb{Q}^s_X)$ be further decreased.
The setting of the hyper-parameters used in this section can found in the Supplementary Materials.

%% file: section/evaluation.tex
\section{Experimental Results}
The supplementary materials can be found from \url{https://arxiv.org/abs/2102.07358}
\subsection{Dataset}
The experiments are conducted on three application scenarios, the digits recognition with domain discrepancy (SVHN\cite{netzer2011reading}, MNIST\cite{deng2012mnist} and USPS\cite{hull1994database} digit datasets), object detection with domain discrepancy (VisDA-C\cite{peng2017visda}), and object detection without domain discrepancy (CIFAR-10\cite{krizhevsky2009learning}). For space, we introduce details of these datasets in the Supplementary Materials.
\subsection{Training Setting}
All experiments are conducted on a server with Ubuntu 18.04 LTS with NVIDIA TITAN RTX GPU cards. The implementation is based on the Pytorch framework.
The hyper-parameter $\alpha$ mentioned above is set to $1e-4$.
We use the standard Adam optimizer~~\cite{kingma2014adam} for optimizing the learning. The network architectures, the learning rate for each part of the network components, the training epoch setting, as well as other hyper-parameters are specified in the Supplementary Materials.
And we get weak annotators in different performance by applying early stop for the training. The implementation details of weak annotators can also be found in the Supplementary Materials.

\subsection{Baseline Experiments Setting}
We conduct comparison experiments with the following baselines. Baseline $B_{wa}$ is the performance of the weak annotator chosen in the experiments in the target domain.
Baseline $B_{t}$ is training $F_1$ only with target data.
Baseline $B_{f_1}$ is a fine-tuning result. It takes the same model as $F_1$ and first uses source domain data and weak labels generated by the weak annotator to train it. Then it uses target domain data to fine-tune the last three layers. 
Baseline $B_{f_2}$ is also a fine-tuning result. The difference is that instead of fine-tuning the last three layers, it trains all network parameters. 

As introduced before, our problem is related to the Semi-Supervised Learning (SSL) and the Semi-Supervised Domain Adaptation (SSDA). For SSL, although we can replace the unlabeled data with samples drawn from another domain instead of the target domain, we cannot find a good way to incorporate the weak annotator into SSL methods for fair comparison with our approach. 
For SSDA, we were able to extend it to our setting for comparison. Specifically, we add 1,000 unlabeled target samples (plus 1,000 labeled target samples, and this setting will be changed accordingly in digits recognition to keep consistent settings) to meet the semi-supervised requirement, and we apply weak annotator to produce weak labels instead of accurate ones for source data. We compare our approach with the following SSDA baselines: FAN~\cite{kim2020attract}, MME~\cite{saito2019semi}, ENT~\cite{grandvalet2004semi}, S+T~\cite{chen2019closer, ranjan2017l2}. Note that to the best of our knowledge, there is no previous work with exactly the same problem setting as ours. The above changes aim at making the comparison as fair as possible. 
Another thing that is worth to mention is that most SSDA methods conduct adaptation on the ImageNet pre-trained models, which introduces a lot of irrelevant data information from the ImageNet dataset. Thus, we disable the pre-training and only allow training with the available data.

\subsection{Results of Digits Recognition}
We evaluate our methods on the digit recognition datasets: SVHN (S), MNIST (M),and USPS (U). 
According to the results shown in Table~\ref{result1}, when the weak annotator performs much worse than the model learned only from the provided target data $B_t$ ($B_{wa}$ = $73.28\%$ on M $\to$ U, $73.28\%$ on S $\to$ U and $76.41\%$ on S $\to$ M), its corresponding baseline $B_{f_1}$ is also lower than $B_t$, and only the second fine-tuning method $B_{f_2}$ is better than or competitive with $B_t$. This indicates that the feature learned from the source domain data and with weak labels introduce data bias, and this bias can be mitigated when the parameters from the front layers are fine-tuned by the target data. 

Overall, we can clearly see that with 15,000 source domain data, limited number of labeled target domain data (second line), and a weak annotator, our method can outperform all the baselines in Table~\ref{result1} with $80.00\%$ on M $\to$ S, $95.99\%$ on M $\to$ U, $96.36\%$ on S $\to$ U and $97.24\%$ on S $\to$ M.

\begin{table} 
\centering
\small
\setlength{\tabcolsep}{1.5mm}{
\begin{tabular*}{0.47\textwidth}{ccccc}  
\hline  

Method & M $\to$ S(\%) & M $\to$ U(\%) & S $\to$ U(\%) & S $\to$ M(\%)\\
\hline
\#samples & 1000 & 300 & 300 & 1000 \\
\hline
$B_{wa}$ & 59.06 & 73.28 & 73.28 & 76.41\\
$B_{t}$ & 61.14 & 89.20 & 89.27 & 94.79\\
$B_{f_1}$ & 55.68 & 84.58 & 77.24 & 80.41\\
$B_{f_2}$ & 77.92 & 94.10 & 94.92 & 95.52\\
S+T & 65.70 & 93.67 & 91.21 & 96.21 \\
ENT & 67.89 & 92.62 & 92.02 & 96.42 \\
MME & 65.92 & 93.07 & 91.32 & 95.64 \\
FAN & 68.48 & 93.78 & 92.38 & 96.51 \\
Ours & \textbf{\color{brown}80.00} & \textbf{\color{brown}95.99} & \textbf{\color{brown}96.36} & \textbf{\color{brown}97.24} \\
\hline  
\end{tabular*}
}
\vspace{0.1cm}
\caption{\small \centering The accuracy of different methods on digit datasets.
}
\vspace{-0.6cm}
\label{result1}
\end{table} 
\subsection{Results of Object Recognition}
\label{obj_reg}
\begin{figure}[htbp]
    \centering
    \includegraphics[width=8.5cm]{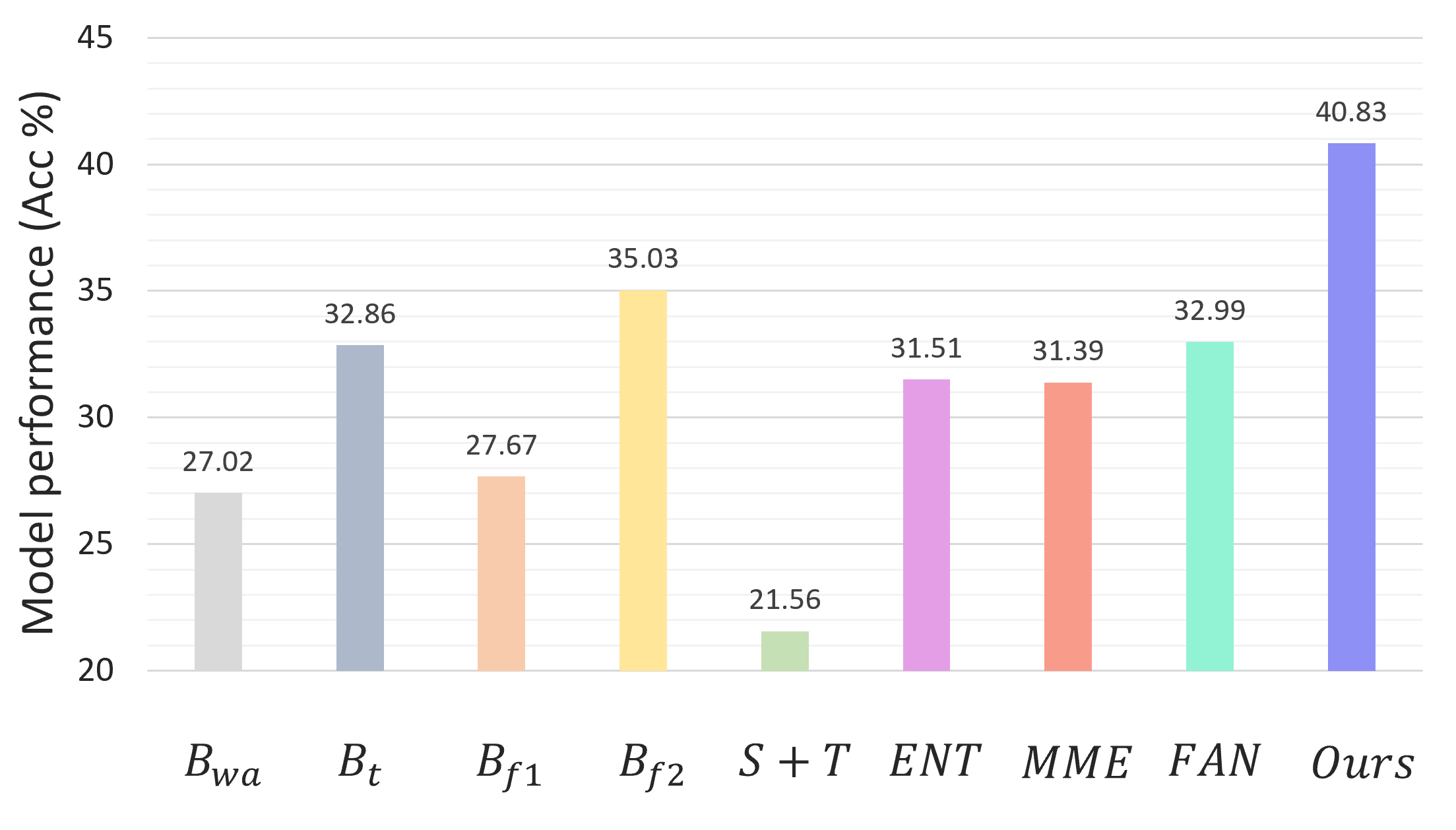}
    \caption{The accuracy of different methods on the VisDA-C dataset. The number is measured in percentage.}
    \label{result2}
\end{figure}

The results of various methods on the VisDA-C dataset are presented in Figure~\ref{result2}. In this task, we utilize the synthetic images as the source domain dataset, and the real-world images as the target domain dataset. And we can see from the table that the performance of the network trained only with the target data is merely $32.86\%$. Then, when the weak annotator is provided, it can help two fine-tune baselines $B_{f_1}$ and $B_{f_2}$ reach $27.67\%$ and $35.03\%$ respectively. As for the SSDA baselines, all of them perform very badly, and they are provided with more target samples with no labels. The best SSDA methods FAN can only achieve $32.99\%$. Our method can provide a result of $40.83\%$, which again exceeds all baselines above. Besides, we also provide additional experiment results using a different weak annotator in the supplementary materials.

Moreover, we also test on the scenario without domain discrepancy using the CIFAR-10 dataset. We randomly select 10,000 data samples from the dataset as the source data and another 1,000 samples as the target data. The result is included in Table~\ref{result3}. As we can see, when the weak annotator is given at $48.96\%$ accuracy, the model trained only with the target data can reach $30.46\%$, while our method nearly doubles the performance and hits $61.71\%$, which exceeds all other baselines.  
\begin{table*} 
\small
\centering
\begin{tabular*}{14.4cm}{cccccccccccccc}  
\hline  
Method & plane & mobile & bird & cat & deer & dog & frog & horse & ship & truck & Acc(\%)\\
\hline
$B_{wa}$ & 43.18 & 65.68 & 28.13 & 25.93 & 29.00 & \textbf{\color{brown}46.15} & \textbf{\color{brown}83.91} & 41.76 & 72.12 & 51.06 & 48.96\\
$B_{t}$ & 19.08 & 63.39 & 03.03 & 30.16 & 25.77 & 22.60 & 46.11 & 50.85 & 23.61 & 25.74 & 30.46\\
$B_{f_1}$ & 57.38 & 77.53 & 38.46 & 33.51 & 45.27 & 33.17 & 73.33 & 58.29 & 57.67 & 60.20 & 52.97\\
$B_{f_2}$ & 44.19 & 80.00 & 38.02 & 41.97 & 47.15 & 24.30 & 78.14 & 55.06 & \textbf{\color{brown}89.35} & 38.19 & 53.49\\
Ours & \textbf{\color{brown}65.52} & \textbf{\color{brown}82.61} & \textbf{\color{brown}39.79} & \textbf{\color{brown}48.45} & \textbf{\color{brown}57.36} & 43.60 & 67.39 & \textbf{\color{brown}65.32} & 70.42 & \textbf{\color{brown}78.89} & \textbf{\color{brown}61.71}\\
\hline  
\end{tabular*}
\caption{The accuracy of different methods on the CIFAR-10 dataset with 10 classes (without domain discrepancy). The number is measured in percentage. The accuracy of each class is from column 2 to column 11. The overall accuracy is shown in the last column.
}
\label{result3}
\end{table*}

\subsection{Ablation Study}
We also study how the quantity of target domain samples and the performance of the weak annotator affect the overall performance of our method. To reduce the impact of domain discrepancy when we study these two factors, we conduct the ablation study on CIFAR-10.
\subsubsection{Sample quantity of target samples}
\begin{figure}[htbp]
    \centering
    \includegraphics[width=8.5cm]{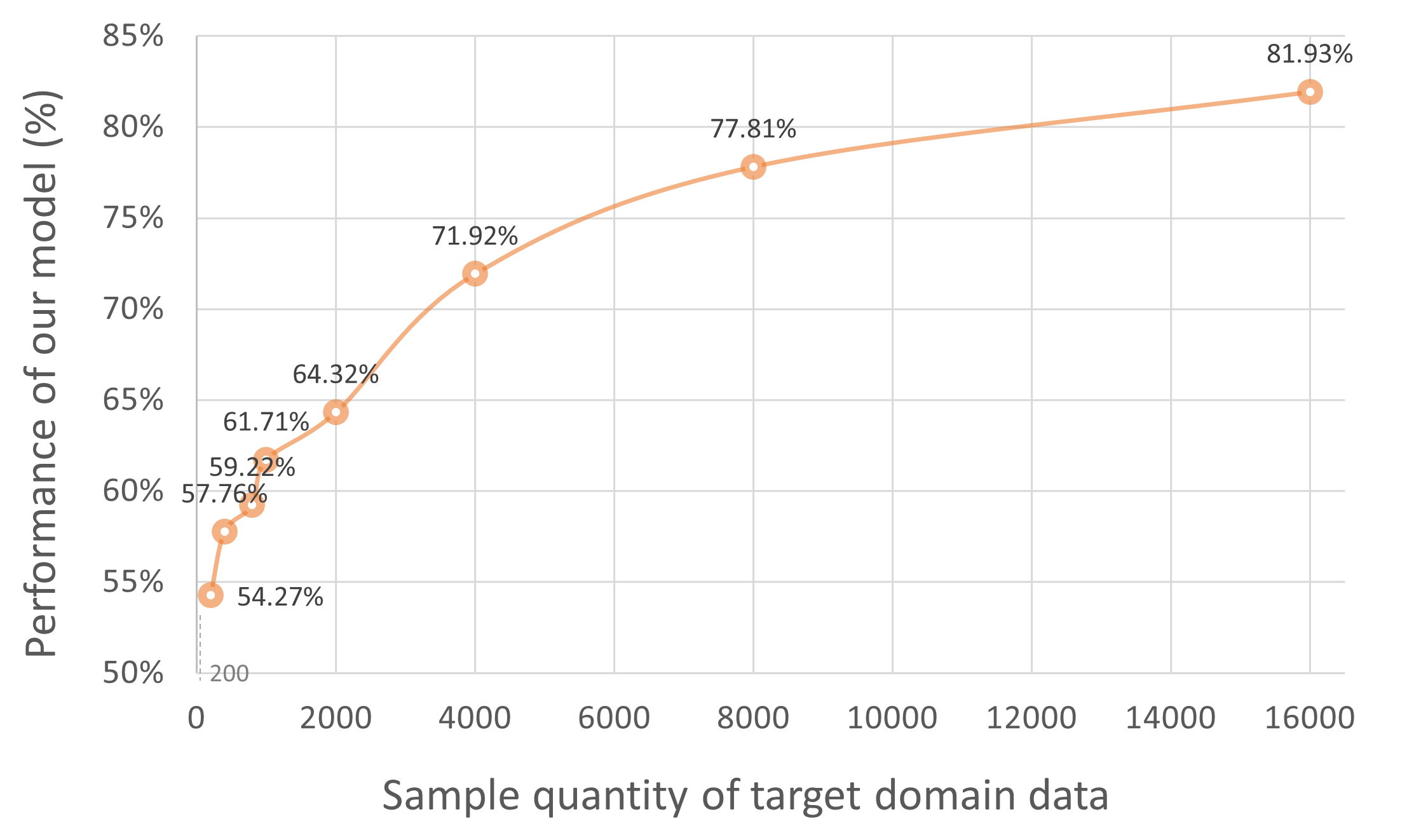}
    \caption{The performance of our learned model under different quantities of target domain samples.}
    \label{figure2}
\end{figure}

As presented in Figure~\ref{figure2}, the horizontal axis indicates the number of target domain data, and the vertical axis shows the performance of our model using the corresponding number of target domain samples. When keeping the weak annotator the same as Section~\ref{obj_reg} and fixing the sample quantity of the source data as 10,000, the accuracy of the model grows as the number of target domain data increases. And it will gradually \textbf{get saturated} when there is enough target domain data. This saturation phenomenon can be explained as the second derivative of $\sqrt{K\!L / N}$ and $\sqrt{\ln{\frac{2N}{\delta}} / N}$ for $N$ is positive while the first derivative is negative. And according to the curve, we can observe that the performance improvement when the target data is less than the source data is relatively higher than the case when there is more target data. The reason for this can be found in our theoretical analysis, i.e., when the sample quantities of source and target data become closer, terms impacted by the quantity of target data will not dominate over the error bound.
    
\subsubsection{Performance of weak annotator}
\begin{figure}[htbp]
    \setlength{\abovecaptionskip}{-0.0cm}
    \setlength{\belowcaptionskip}{-0.1cm}
    \centering
    \includegraphics[width=8.5cm]{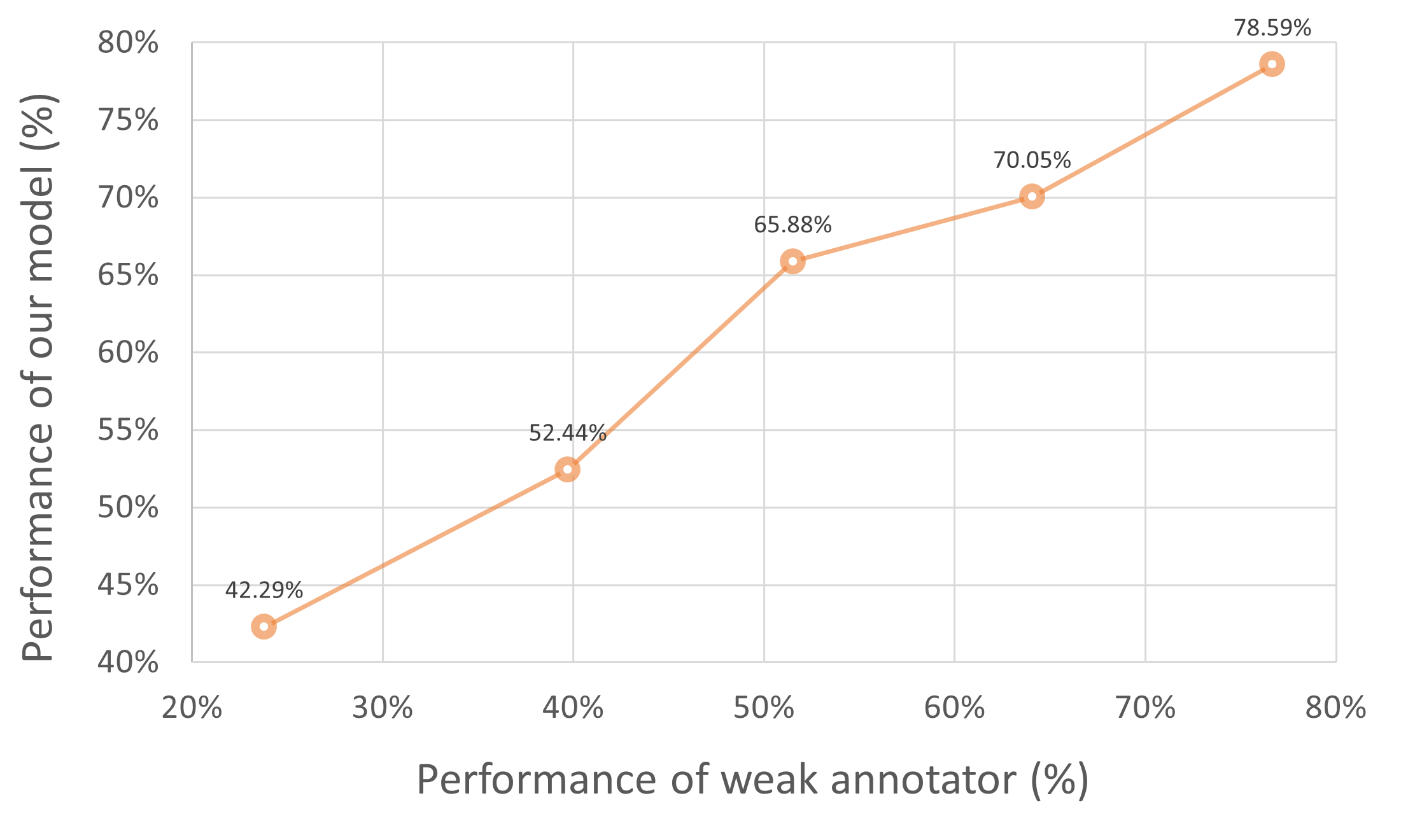}
    \caption{The performance of our learned model under different accuracy of the weak annotator.}
    \label{figure3}
\end{figure}

Figure~\ref{figure3} shows the curve of how the performance of our model changes with respect to the accuracy of the weak annotator. As shown in the figure, when the weak annotator performs the worst with accuracy of $23.79\%$, our model can reach $42.29\%$, which is a relatively significant improvement. And as the accuracy of the weak annotator increases, our model performs better accordingly. Interestingly, the improvement curve in Figure~\ref{figure3} is approximately linear, which demonstrates that it is reasonable to \textbf{linearly} add the terms of the weak annotator in the error bound. 

%% file: section/conclusion.tex
\section{Conclusion} \label{sec:conclusion}
In this work, we present a novel approach leveraging weak annotator to address the data insufficiency challenge in domain adaptation, where only a small amount of data samples is available in the target domain and the data samples in the source domain are unlabeled. 
Our weak adaptation approach includes a theoretical analysis that derives the error bound of a trained classifier with respect to the data quantity and the performance of the weak annotator, and a multi-stage learning process that improves classifier performance by lowering the error bound. 
Our approach shows significant improvement over baselines on cases with or without domain discrepancy in various data sets.


%% file: section/acknowledgement.tex
\section{Acknowledgement}

We gratefully acknowledge the support from NSF grants 1834701, 1839511, 1724341, 2038853, and ONR grant N00014-19-1-2496.

%% file: section/Appendix.tex
This appendix contains additional details for the submitted article \textbf{\emph{Weak Adaptation Learning: Addressing Cross-domain Data Insufficiency with Weak Annotator}}, including mathematical proofs and experimental details. Moreover, the additional related works are provided in Section~\ref{Appendix-related_work}, the additional theoretical analysis details are provided in Section~\ref{Appendix-theory}, the introduction of data sets is provided in Section~\ref{Appendix-dataset}, the additional ablation study is provided in Section~\ref{Appendix-ab}, the network architecture is shown in Section~\ref{Appendix-Network-arch}, and more about training settings is presented in Section~\ref{Appendix-train}.

\section{Related Work}
\label{Appendix-related_work}
\subsection{Weakly- and Semi-Supervised Learning}%

Weakly Supervised Learning is a large concept that may have multiple problem settings~\cite{zhou2018brief}. The problem we consider in this paper is related to the incomplete supervision setting that is often addressed by Semi-Supervised Learning (SSL) approaches.
Standard SSL solves the problem of training a model with a few labeled data and a large amount of unlabeled data. Some of the widely-applied methods~\cite{rasmus2015semi, dong2018tri, perez2018weakly, belharbi2020deep} assign pseudo labels to unlabeled samples and then perform supervised learning. And there are works that address the noises in the labels of those samples~\cite{natarajan2013learning, ghosh2015making, liu2015classification}. 
Our target problem is related to SSL with inaccurate supervision, but is different since we consider the feature discrepancy between the (unlabeled) source data and the (labeled) target data -- a case that occurs often in practice but has not been sufficiently addressed.  

Positive-unlabeled Learning (PuL) is usually regarded as a sub-problem of SSL. Its goal is to learn a binary classifier to distinguish positive and negative samples from a large amount of unlabeled data and a few positive samples. Several works~\cite{kiryo2017positive, chen2020self} can achieve great performance on selecting samples that are similar to the positive data, and there are also works using samples selected by PuL to perform other tasks~\cite{xu2019positive, loghmani2020positive}.

\subsection{Domain Adaptation}
Previous Domain Adaptation (DA) approaches often design a number of losses to measure the discrepancy between source and target domains~\cite{saito2018maximum, long2016unsupervised, tzeng2014deep, zellinger2017central, roy2019unsupervised}. Then, adversarial training is applied in DA~\cite{mathieu2016disentangling, huang2018auggan, shermin2020adversarial, tang2020discriminative}. In the problem settings of these works, the supervision knowledge reflected in the source labels is accurate and reliable, which is not the case in our problem (where weak annotator generates inaccurate labels). 
In addition, the standard DA problem does not address the case where the quantity of the target domain data is small.  
The Semi-Supervised Domain Adaptation (SSDA)~\cite{saito2019semi} approach assumes that only a few labeled data samples are available in the target domain, but the adaptation may use a large amount of unlabeled data in the target domain.  
This is different from our problem, where we consider the cases where even unlabeled data samples are hard to acquire in the target domain (e.g., the examples mentioned in Section~\ref{sec:introduction}).
In short, our work is the first to explore the DA problem with weak source knowledge and insufficient target samples simultaneously.

In addition, there are several papers in domain adaptation that leverage the information of so-called \textit{weak labels} in source or target domain~\cite{wilson2020multi, paul2020domain}. However, these weak labels are in a different task to their target labels (e.g., providing an image-level
label for image segmentation task that shows the proportion of categories of the entire image rather than providing a pixel-level label), which is a different concept towards the weak label mentioned in our paper.

\subsection{Importance of Sample Quantity}
The training of machine learning models, especially deep neural networks, often requires a large amount of data samples. However, in many practical scenarios, there is not sufficient training data to feed the learning process, degrading the model performance sharply~\cite{snell2017prototypical, johnson2019survey}. Many approaches have been proposed to make up for the lack of training samples, e.g., data re-sampling~\cite{van2007experimental}, data augmentation~\cite{pouyanfar2018dynamic}, metric learning, and meta learning~\cite{cao2019theoretical, chen2019closer, triantafillou2019meta}. And there are works~\cite{neyshabur2017exploring, amit2018meta, cao2019theoretical, wang2020generalizing} conducting theoretical analysis on the relation between training data quantity and model performance. These analyses are usually in the form of bounding the prediction error of the models and provide valuable information on how the sample quantity of training data affects the model performance. In our work, we also perform a theoretical analysis on the error bound of the trained model, with respect to not only the data quantity but also the performance of the weak annotator.

\section{Theoretical Analysis}
\label{Appendix-theory}
\subsection{Proof of Theorem 1}
\noindent \emph{\textbf{Theorem 1.}} \textit{For a classifier parameter distribution $\mathcal{H} \sim \mathcal{N}(0, \sigma_{\mathcal{H}}^2)$ that is independent of the training data with size $m$, and a posterior parameter distribution $\mathcal{Q}$ learned from the training data set, if we assume $\mathcal{Q} \sim \mathcal{N}(\mu_{\mathcal{Q}}, \sigma_{\mathcal{Q}}^2)$ and consider $\Theta^{\mathbf{p}}$, $\Theta$ as drawn from $\mathcal{H}$, $\mathcal{Q}$ respectively ($\Theta = \Theta^{\mathbf{p}} + \nabla (\widehat{L}(\mathbf{h}_{\Theta^{\mathbf{p}}}))$), the KL divergence of $\mathcal{Q}$ and $\mathcal{H}$ is bounded with symbols defined in Eq.~(\ref{distribution of learned parameter}) as follows:}
\begin{equation}
    K\!L(\mathcal{Q}\, \|\, \mathcal{H}) \le \frac{\frac{\sigma^2}{m} + \mu^2}{2\sigma_{\mathcal{H}}^2}
\end{equation}
\textbf{\emph{Proof:}} Let us consider a classifier $\mathbf{h}_{\Theta}$ drawn from a parameter distribution $\mathcal{Q}$, and $\mathcal{Q}$ depends on the data that has been learned by $\mathbf{h}$. Then, if we denote the model parameters of $\mathbf{h}$ before the training that are drawn from $\mathcal{H}$ as $\Theta^{\mathbf{p}}$, the KL divergence can be written as $K\!L(\mathcal{Q} \| \mathcal{H}) = -\mathbb{E} [ \Theta \cdot (\ln{\Theta^{\mathbf{p}}} - \ln{\Theta}) ]$. As aforementioned, $\mathbf{h}_{\Theta}$ is trained with the training data set from $\mathbf{h}_{\Theta^{\mathbf{p}}}$, and we consider that the training is optimized by a gradient-based method. Thus, we can formulate that $\Theta = \Theta^{\mathbf{p}} + \nabla (\widehat{L}(\mathbf{h}_{\Theta^{\mathbf{p}}}))$. Here we omit the learning rate to simplify the formula, and $\widehat{L}(\mathbf{h}_{\Theta})$ is the empirical error computed from the training set ($\widehat{L}(\mathbf{h}_{\Theta}) = \frac{1}{m} \sum_{i=1}^m \mathcal{L}(\mathbf{x}_i, y_i)$, where $\mathcal{L}$ denotes the loss of a single training sample).

The PAC-Bayesian error bound is valid for any parameter distribution $\mathcal{H}$ that is independent of the training data, and any method of optimizing $\Theta^{\mathbf{p}}$ dependent on the training set~\cite{neyshabur2017exploring}. Therefore, in order to simplify the problem, we instantiate the bound as setting $\mathcal{H}$ to conform to a Gaussian distribution with zero mean ($\mu_{\mathcal{H}} = 0$) and $\text{Var}_{\mathcal{H}} = \sigma^2_{\mathcal{H}}$ variance. This simplification is similar to the ones in previous PAC-Bayesian works~\cite{neyshabur2017exploring, neyshabur2018pac}. We further assume that the parameter change of the overall model during training can also be regarded as conforming to an empirical Gaussian distribution, and we denote this distribution as follows:
\begin{equation}
\begin{aligned}
    \mu &\triangleq \mathbb{E} \left[\nabla_{\Theta^{\mathbf{p}}} \mathcal{L}(\mathbf{x}, y)\right] \\
    \sigma^2 &\triangleq \mathbb{E} \left[(\nabla_{\Theta^{\mathbf{p}}} \mathcal{L}(\mathbf{x}, y) - \mu)(\nabla_{\Theta^{\mathbf{p}}} \mathcal{L}(\mathbf{x}, y) - \mu)^T\right]
\end{aligned}
\label{distribution of learned parameter}
\end{equation}
This Gaussian distribution is independent of model parameters if we regard the parameter updates induced by the gradient back-propagation as accumulated random perturbations, i.e., each training sample corresponds to a small perturbation~\cite{neyshabur2018pac}. And we know that the specific formula of KL divergence to any two Gaussian distributions $p \sim \mathcal{N}(\mu_1, \sigma_1^2)$, $q \sim \mathcal{N}(\mu_2, \sigma_2^2)$ is written as:
\begin{equation}
    K\!L(p, q) = \ln{\frac{\sigma_2}{\sigma_1}} + \frac{\sigma_1^2 + (\mu_1 - \mu_2)^2}{2\sigma_2^2} - \frac{1}{2}
\end{equation}
And we can prove the equation as follows:

\begin{equation}
\begin{aligned}
K\!L(p, q) &=\int[\log (p(x))-\log (q(x))] p(x) d x \\
&=\int[p(x) \log (p(x))-p(x) \log (q(x))] d x
\end{aligned}
\label{KL_relation}
\end{equation}

Then we can split Eq.~(\ref{KL_relation}) into two different integrations and calculate them respectively.

\begin{equation}
\small
\begin{aligned}
&\int p(x) \log (p(x)) d x \\
&=\int p(x) \log \left[\frac{1}{\sqrt{2 \pi} \sigma_{1}} \exp \left(-\frac{\left(x-\mu_{1}\right)^{2}}{2 \sigma_{1}^{2}}\right)\right] d x \\
&=\int p(x)\left[\log \frac{1}{\sqrt{2 \pi} \sigma_{1}}+\log \exp \left(-\frac{\left(x-\mu_{1}\right)^{2}}{2 \sigma_{1}^{2}}\right)\right] d x \\
&=-\frac{1}{2} \log \left(2 \pi \sigma_{1}^{2}\right)+\int p(x)\left(-\frac{\left(x-\mu_{1}\right)^{2}}{2 \sigma_{1}^{2}}\right) d x \\
&=-\frac{1}{2} \log \left(2 \pi \sigma_{1}^{2}\right)-\frac{\left(\mu_{1}^{2}+\sigma_{1}^{2}\right)-\left(2 \mu_{1} \cdot \mu_{1}\right)+\mu_{1}^{2}}{2 \sigma_{1}^{2}} \\
&=-\frac{1}{2}\left[1+\log \left(2 \pi \sigma_{1}^{2}\right)\right]
\end{aligned}
\end{equation}

\begin{equation}
\small
\begin{aligned}
&\int p(x) \log (q(x)) d x \\ 
&=\int p(x) \log \left[\frac{1}{\sqrt{2 \pi} \sigma_{2}} \exp \left(-\frac{\left(x-\mu_{2}\right)^{2}}{2 \sigma_{2}^{2}}\right)\right] d x \\
&=\int p(x)\left[\log \frac{1}{\sqrt{2 \pi} \sigma_{2}}+\log \exp \left(-\frac{\left(x-\mu_{2}\right)^{2}}{2 \sigma_{2}^{2}}\right)\right] d x \\
&=-\frac{1}{2} \log \left(2 \pi \sigma_{2}^{2}\right)+\int p(x)\left(-\frac{\left(x-\mu_{2}\right)^{2}}{2 \sigma_{2}^{2}}\right) d x \\
&=\frac{1}{2} \log \left(2 \pi \sigma_{2}^{2}\right)-\frac{\left(\mu_{1}^{2}+\sigma_{1}^{2}\right)-\left(2 \mu_{2} \cdot \mu_{1}\right)+\mu_{2}^{2}}{2 \sigma_{2}^{2}} \\
&=-\frac{1}{2} \log \left(2 \pi \sigma_{2}^{2}\right)-\frac{\sigma_{1}^{2}+\left(\mu_{1}-\mu_{2}\right)^{2}}{2 \sigma_{2}^{2}}
\end{aligned}
\end{equation}

In this case, Eq.~(\ref{KL_relation}) can be written as:
\begin{equation}
\small
\begin{aligned}
&K\!L(p, q) =\int[p(x) \log (p(x))-p(x) \log (q(x))] d x \\
&=-\frac{1}{2}\left[1+\log \left(2 \pi \sigma_{1}^{2}\right)\right]+\frac{1}{2} \log \left(2 \pi \sigma_{2}^{2}\right)+\frac{\sigma_{1}^{2}+\left(\mu_{1}-\mu_{2}\right)^{2}}{2 \sigma_{2}^{2}} \\
&=\log \frac{\sigma_{2}}{\sigma_{1}}+\frac{\sigma_{1}^{2}+\left(\mu_{1}-\mu_{2}\right)^{2}}{2 \sigma_{2}^{2}}-\frac{1}{2}
\end{aligned}
\end{equation}

With the above definitions, we can compute the mean of distribution $\mathcal{Q}$:
\begin{equation}
\begin{aligned}
    \mu_{\mathcal{Q}} &= \mathbb{E}\left \{\Theta^{\mathbf{p}} + \nabla[\widehat{L}(\mathbf{x}, y)]\right \}\\
    &= \mu_{\mathcal{H}} + \mathbb{E}\left\{\nabla\left[\frac{1}{m} \sum_{j=1}^{m}\mathcal{L}(\mathbf{x}_j, y_j)\right]\right\}\\
    &= 0 + \frac{1}{m}\sum_{j=1}^{m}\mathbb{E}\left[\nabla\mathcal{L}(\mathbf{x}_j, y_j)\right]\\
    &= \mu
    \end{aligned}
    \end{equation}
\begin{equation}
\small
\begin{aligned}
    \sigma_{\mathcal{Q}}^2 &= {V\!ar}\left \{\Theta^{\mathbf{p}} + \nabla[\widehat{L}(\mathbf{x}, y)]\right \}\\
    &= {V\!ar}(\Theta^{\mathbf{p}}) + {V\!ar}\left\{\nabla\left[\frac{1}{m} \sum_{j=1}^{m}\mathcal{L}(\mathbf{x}_j, y_j)\right]\right\}\\
    &= \sigma_{\mathcal{H}}^2 + \frac{1}{m^2}{V\!ar}\left\{\sum_{j=1}^m\nabla[\mathcal{L}(\mathbf{x}_j, y_j)]\right\}\\
    &= \sigma_{\mathcal{H}}^2 + \frac{1}{m^2}\sum_{j=1}^m {V\!ar}\left\{\nabla[\mathcal{L}(\mathbf{x}_j, y_j)]\right\}\\
    &= \sigma_{\mathcal{H}}^2 + \frac{1}{m} \sigma^2
\end{aligned}
\end{equation}

With the mean and variance of distribution $\mathcal{Q}$, we can then calculate $K\!L(\mathcal{Q}\, \|\, \mathcal{H})$ as follows:
\begin{equation}
\begin{aligned}
    K\!L(\mathcal{Q} \| \mathcal{H}) &= \ln{\frac{\sigma_{\mathcal{H}}}{\sigma_{\mathcal{Q}}}} + \frac{\sigma_{\mathcal{Q}}^2 + (\mu_{\mathcal{Q}} - \mu_{\mathcal{H}})^2}{2\sigma_{\mathcal{H}}^2} - \frac{1}{2}\\
    &\leq \frac{\sigma_{\mathcal{H}}^2 
    + \frac{1}{m}\sigma^2 + \mu^2}{2\sigma_{\mathcal{H}}^2} - \frac{1}{2}\\
    &\quad = \frac{\frac{1}{m}\sigma^2 + \mu^2}{2\sigma_{\mathcal{H}}^2}
\end{aligned}
\end{equation}
$\hfill \blacksquare$

\subsection{Error Bound Derivation}
With the risk definition introduced in our main paper, the risk of $\mathbf{h}$ with respect to source data distribution $\mathbb{Q}^s$ is
\begin{equation}
\centering
R^s(\mathbf{h}) = \mathbb{E}_{(\mathbf{x},y) \sim \mathbb{Q}^s} \mathcal{L}(\mathbf{h}(\mathbf{x}), y) = L(\mathbf{h}_{\Theta})_{\sim \mathbb{Q}^s}
\end{equation}
The risk with respect to target data distribution $\mathbb{Q}^t$ is 
\begin{equation}
\setlength{\abovedisplayskip}{3pt}
\setlength{\belowdisplayskip}{3pt}
\centering
R^t(\mathbf{h}) = \mathbb{E}_{(\mathbf{x},y) \sim \mathbb{Q}^t} \mathcal{L}(\mathbf{h}(\mathbf{x}), y) = L(\mathbf{h}_{\Theta})_{\sim \mathbb{Q}^t}
\end{equation}
In addition, we define the Classification Distance of two classifier $\mathbf{h}_1$ and $\mathbf{h}_2$ under a same domain distribution $\mathbb{P}$ as 
\begin{equation}
\mathcal{CD}_{\sim \mathbb{P}}(\mathbf{h}_1, \mathbf{h}_2) = \mathbb{E}_{\mathbf{x} \sim \mathbb{P}} \mathcal{L}(\mathbf{h}_1(\mathbf{x}), \mathbf{h}_2(\mathbf{x}))
\end{equation}
Moreover, the Discrepancy Distance of two domains is defined as in~\cite{mansour2009domain}:  $\forall \mathbf{h_1}, \mathbf{h_2}$, the discrepancy distance between the distributions of two domains $\mathbb{P}, \mathbb{Q}$ is
\begin{equation}
    \setlength{\abovedisplayskip}{3pt}
    \setlength{\belowdisplayskip}{3pt}
    \mathcal{DD} (\mathbb{P}, \mathbb{Q}) = \sup_{\mathbf{h_1}, \mathbf{h_2} \in \mathbb{H}} |\mathcal{CD}_{\sim \mathbb{P}}(\mathbf{h}_1, \mathbf{h}_2) - \mathcal{CD}_{\sim \mathbb{Q}}(\mathbf{h}_1, \mathbf{h}_2)|
    \label{domain discrepancy}
\end{equation}

With the operators $\oplus$ and $\ominus$ defined in our main paper, we can obtain the following theorem for error bound with all L1, L2 and their combination loss functions.

\noindent \emph{\textbf{Theorem 2.}} \textit{For all L1 (Mean Absolute Error~\cite{kassam1978quantization}), L2 (Mean Squared Error~\cite{gunst1977biased}) and their non-negative combination loss functions (Huber Loss~\cite{yi2017semismooth}, Quantile Loss~\cite{koenker2004quantile}, etc.), the classification risk of aforementioned $\mathbf{h} \oplus \mathbf{d}$ can be formulated as follows:}
\begin{equation}
\begin{aligned}
    R^t(\mathbf{h} \oplus \mathbf{d})
    &= \mathbb{E}_{\mathbb{Q}^t_X}\mathcal{L}(\mathbf{h} \oplus \mathbf{d}, \mathbf{h}^{w} \oplus \mathbf{h}^{o_t} \ominus \mathbf{h}^w)\\
    &\leq \mathbb{E}_{\mathbb{Q}^t_X}\mathcal{L}(\mathbf{h}, \mathbf{h}^{w}) + \mathbb{E}_{\mathbb{Q}^t_X}\mathcal{L}(\mathbf{d}, \mathbf{h}^{o_t} \ominus \mathbf{h}^w)\\
\end{aligned}
\label{basic term of overall bound}
\end{equation}
\textbf{\emph{Proof:}} For L1 (Mean Absolute Error) Loss, $\mathcal{L}(\mathbf{h}_1, \mathbf{h}_2) = |\mathbf{h}_1(\mathbf{x}) - \mathbf{h}_2(\mathbf{x})|$. Then, incorporating the definitions of new operators, we can have the following derivation (here we denote $\mathcal{L}(\mathbf{h} \oplus \mathbf{d}, \mathbf{h}^{w} \oplus \mathbf{h}^{o_t} \ominus \mathbf{h}^w)$ as $\mathcal{L}$):
\begin{equation}
\begin{aligned}
     \mathcal{L} &= \left|[\mathbf{h}(\mathbf{x}) + \mathbf{d}(\mathbf{x})] - [\mathbf{h}^{w}(\mathbf{x}) + \mathbf{h}^{o_t}(\mathbf{x}) - \mathbf{h}^{w}(\mathbf{x})]\right|\\
     &= \left|[\mathbf{h}(\mathbf{x}) - \mathbf{h}^{w}(\mathbf{x})] + [\mathbf{d}(\mathbf{x}) -  \mathbf{h}^{o_t}(\mathbf{x}) + \mathbf{h}^{w}(\mathbf{x})]\right|\\
     &\leq |\mathbf{h}(\mathbf{x}) - \mathbf{h}^{w}(\mathbf{x})| + |\mathbf{d}(\mathbf{x}) -  \mathbf{h}^{o_t}(\mathbf{x}) + \mathbf{h}^{w}(\mathbf{x})|\\
     &\quad = \mathcal{L}(\mathbf{h}, \mathbf{h}^{w}) + \mathcal{L}(\mathbf{d}, \mathbf{h}^{o_t} \ominus \mathbf{h}^w)
\end{aligned}
\end{equation}
For L2 (Mean Squared Error) Loss, $\mathcal{L}(\mathbf{h}_1, \mathbf{h}_2) = (\mathbf{h}_1(\mathbf{x}) - \mathbf{h}_2(\mathbf{x}))^2$, and the corresponding bound of $\mathcal{L}(\mathbf{h} \oplus \mathbf{d}, \mathbf{h}^{w} \oplus \mathbf{h}^{o_t} \ominus \mathbf{h}^w)$ is as follows:
\begin{equation}
\begin{aligned}
     \mathcal{L} &= \left\{[\mathbf{h}(\mathbf{x}) + \mathbf{d}(\mathbf{x})] - [\mathbf{h}^{w}(\mathbf{x}) + \mathbf{h}^{o_t}(\mathbf{x}) - \mathbf{h}^{w}(\mathbf{x})]\right\}^2\\
     &= \left\{[\mathbf{h}(\mathbf{x}) - \mathbf{h}^{w}(\mathbf{x})] + [\mathbf{d}(\mathbf{x}) -  \mathbf{h}^{o_t}(\mathbf{x}) + \mathbf{h}^{w}(\mathbf{x})]\right\}^2
\end{aligned}
\label{MSE bound 1}
\end{equation}
Here, we denote $[\mathbf{h}(\mathbf{x}) - \mathbf{h}^{w}(\mathbf{x})]$ as $t_1$, and $[\mathbf{d}(\mathbf{x}) -  \mathbf{h}^{o_t}(\mathbf{x}) + \mathbf{h}^{w}(\mathbf{x})]$ as $t_2$. As aforementioned, $\mathbf{d}$ is designed to approximate the difference between the weak supervision $\mathbf{h}$ and the ground truth $\mathbf{h}^{o_t}$ (in our method, we show that $\mathbf{h}$ can be directly replaced with $\mathbf{h}^w$), and thus we assume that the output of $\mathbf{d}$ is very closed to $\mathbf{h}^{o_t} - \mathbf{h}$. Then, we analyze the relation between the product of $t_1, t_2$ and $0$ in all 6 possible cases shown in Table~\ref{MSE derivation}.

\begin{table}[htbp]
\centering
\begin{tabular}{ccc}
\hline
\textbf{Condition} & \textbf{Derivation} & \textbf{Conclusion} \\ \hline
$\mathbf{h}^{o_t} \geq \mathbf{h} \geq \mathbf{h}^w$           &          $\mathbf{h}^{o_t} - \mathbf{h} \leq \mathbf{h}^{o_t} - \mathbf{h}^w$           &           $t_1 \times t_2 \leq 0$          \\
$\mathbf{h}^{o_t} \geq \mathbf{h}^w \geq \mathbf{h}$           &          $\mathbf{h}^{o_t} - \mathbf{h} \geq \mathbf{h}^{o_t} - \mathbf{h}^w$           &           $t_1 \times t_2 \leq 0$          \\
$\mathbf{h} \geq \mathbf{h}^{o_t} \geq \mathbf{h}^w$           &          $\mathbf{h}^{o_t} - \mathbf{h} \leq \mathbf{h}^{o_t} - \mathbf{h}^w$           &           $t_1 \times t_2 \leq 0$          \\
$\mathbf{h} \geq \mathbf{h}^w \geq \mathbf{h}^{o_t}$           &          $\mathbf{h}^{o_t} - \mathbf{h} \leq \mathbf{h}^{o_t} - \mathbf{h}^w$           &           $t_1 \times t_2 \leq 0$          \\
$\mathbf{h}^w \geq \mathbf{h}^{o_t} \geq \mathbf{h}$           &          $\mathbf{h}^{o_t} - \mathbf{h} \geq \mathbf{h}^{o_t} - \mathbf{h}^w$           &           $t_1 \times t_2 \leq 0$          \\
$\mathbf{h}^w \geq \mathbf{h} \geq \mathbf{h}^{o_t}$           &          $\mathbf{h}^{o_t} - \mathbf{h} \geq \mathbf{h}^{o_t} - \mathbf{h}^w$           &           $t_1 \times t_2 \leq 0$          \\ \hline
\end{tabular}
\caption{Compare the product of $[\mathbf{h}(\mathbf{x}) - \mathbf{h}^{w}(\mathbf{x})]$, $[\mathbf{d}(\mathbf{x}) -  \mathbf{h}^{o_t}(\mathbf{x}) + \mathbf{h}^{w}(\mathbf{x})]$ with $0$ in all 6 possible cases.}
\label{MSE derivation}
\end{table}
According to Table~\ref{MSE derivation}, the product of $t_1, t_2$ is always less than or equal to $0$. Therefore, we can enlarge the bound of Eq.~({\ref{MSE bound 1}}) as follows:
\begin{equation}
\begin{aligned}
     \mathcal{L} &\leq \{\mathbf{h}(\mathbf{x}) - \mathbf{h}^{w}(\mathbf{x})\}^2 + \{\mathbf{d}(\mathbf{x}) -  \mathbf{h}^{o_t}(\mathbf{x}) + \mathbf{h}^{w}(\mathbf{x})\}^2\\
     &\quad = \mathcal{L}(\mathbf{h}, \mathbf{h}^{w}) + \mathcal{L}(\mathbf{d}, \mathbf{h}^{o_t} \ominus \mathbf{h}^w)
\end{aligned}
\end{equation}
Now, we have proven that both L1 and L2 Losses hold true for Eq.~(\ref{basic term of overall bound}). Moreover, we believe that other losses that are non-negative combination forms of L1 and L2 can also ensure this inequality relation.
$\hfill \blacksquare$

From the definition of Discrepancy Distance between two domains shown in Eq.~(\ref{domain discrepancy}), we can derive the inequality relation on the discrepancy between $\mathbb{Q}_X^t$ and $\mathbb{Q}_X^s$ as:
\begin{equation}
    \mathbb{E}_{\mathbb{Q}^t_X}\mathcal{L}(\mathbf{h}, \mathbf{h}^{w}) \leq \mathcal{DD}(\mathbb{Q}_X^t, \mathbb{Q}_X^s) + \mathbb{E}_{\mathbb{Q}^w_X}\mathcal{L}(\mathbf{h}, \mathbf{h}^{w})
    \label{domain discrepancy inequality}
\end{equation}
By utilizing the triangle inequality, we have:
\begin{equation}
\begin{aligned}
    \mathbb{E}_{\mathbb{Q}^t_X}\mathcal{L}(\mathbf{h}, \mathbf{h}^{w}) 
    &\leq \mathbb{E}_{\mathbb{Q}^t_X}\mathcal{L}(\mathbf{h}, \mathbf{h}^{o_t}) + \mathbb{E}_{\mathbb{Q}^t_X}\mathcal{L}(\mathbf{h}^{o_t}, \mathbf{h}^{w}) \\
    &\leq R^t(\mathbf{h}) + R^t(\mathbf{h}^{w})
    \label{triangle inequality 1}
\end{aligned}
\end{equation}
\begin{equation}
\begin{aligned}
    R^t(\mathbf{h}) 
    &\leq \mathbb{E}_{\mathcal{Q}^t_X}\mathcal{L}(\mathbf{h}, \mathbf{h}^{w}) + \mathbb{E}_{\mathbb{Q}^t_X}\mathcal{L}(\mathbf{h}^{w}, \mathbf{h}^{o_t})\\
    &\leq \mathcal{CD}_t(\mathbf{h}, \mathbf{h}^{w}) + R^t(\mathbf{h}^{w} )
    \label{triangle inequality 2}
\end{aligned}
\end{equation}
Based on the relations presented in Eq.~(\ref{domain discrepancy inequality}), (\ref{triangle inequality 1}), (\ref{triangle inequality 2}), we can bound the classification risk of $\mathbf{h} \oplus \mathbf{d}$ in the target domain shown in Eq.~(\ref{basic term of overall bound}) as: 
\begin{equation}
\small
\begin{aligned}
    &R^t(\mathbf{h} \oplus \mathbf{d}) \\
    &\leq \mathbb{E}_{\mathcal{Q}^t_X}\mathcal{L}(\mathbf{h}, \mathbf{h}^{w}) + \mathbb{E}_{\mathcal{Q}^t_X}\mathcal{L}(\mathbf{d}, \mathbf{h}^{o_t} \ominus \mathbf{h}^w)\\
    &\leq R^t(\mathbf{h}) + R^t(\mathbf{h}^{w}) + \mathcal{CD}_t(\mathbf{d}, \mathbf{h}^{o_t} \ominus \mathbf{h}^w)\\
    &\leq R^t(\mathbf{h}^{w} ) + \mathcal{CD}_t(\mathbf{h}, \mathbf{h}^{w}) + R^t(\mathbf{h}^{w}) + \mathcal{CD}_t(\mathbf{d}, \mathbf{h}^{o_t} \ominus \mathbf{h}^w)\\
    &\leq 2R^t(\mathbf{h}^{w} ) + \mathcal{CD}_s(\mathbf{h}, \mathbf{h}^{w}) + \mathcal{DD}(\mathbb{Q}^t_X, \mathbb{Q}^s_X)\\
    &\quad+ \mathcal{CD}_t(\mathbf{d}, \mathbf{h}^{o_t} \ominus \mathbf{h}^w)
\end{aligned}
\label{second version of overall bound}
\end{equation}
Furthermore, the Classification Discrepancy between $\mathbf{h}$ and $\mathbf{h}^w$ for the source domain data can also be bounded by the triangle inequality:
\begin{equation}
\begin{aligned}
    \mathbb{E}_{\mathcal{Q}^s_X}\mathcal{L}(\mathbf{h}, \mathbf{h}^{w}) 
    &\leq \mathbb{E}_{\mathcal{Q}^s_X}\mathcal{L}(\mathbf{h}, \mathbf{h}^{o_s}) + \mathbb{E}_{\mathcal{Q}^s_X}\mathcal{L}(\mathbf{h}^{o_s}, \mathbf{h}^{w}) \\
    \mathcal{CD}_s(\mathbf{h}, \mathbf{h}^w) &\leq R^s(\mathbf{h}) + R^s(\mathbf{h}^{w})
\end{aligned}
\end{equation}

Thus, Eq.~(\ref{second version of overall bound}) can be formulated as follows:
\begin{equation}
\begin{aligned}
    R^t(\mathbf{h} \oplus \mathbf{d}) &\leq 2R^t(\mathbf{h}^{w} ) + R^s(\mathbf{h}) + R^s(\mathbf{h}^{w})\\ &\quad+ \mathcal{DD}(\mathbb{Q}^t_X, \mathbb{Q}^s_X) + \mathcal{CD}_t(\mathbf{d}, \mathbf{h}^{o_t} \ominus \mathbf{h}^w)
    \label{basic term of overall bound 1}
\end{aligned}
\end{equation}
As introduced in our main paper, if $\mathbf{h}_{\Theta}$ is trained with mini-batch, its classification error can be bounded as follows~\cite{neyshabur2017exploring}:
\begin{equation}
\begin{aligned}
    L(\mathbf{h}_{\Theta}) &\leq \widehat{L}(\mathbf{h}_{\Theta})+4\sqrt{\frac{\left(K\!L(\mathcal{Q} \| \mathcal{H})+\ln \frac{2m}{\delta}\right)}{m}}\\
    &\leq \widehat{L}(\mathbf{h}_{\Theta}) + 4\sqrt{\frac{\left(K\!L(\mathcal{Q} \| \mathcal{H})\right)}{m}} + 4\sqrt{\frac{\ln{\frac{2m}{\delta}}}{m}}
\end{aligned}
\label{pac_theorem_further}
\end{equation}
In our setting, the weak annotator is unchanged. Thus the KL divergence of the risk for $\mathbf{h}^w$ is zero. As mentioned before, classifier $\mathbf{d}$ is designed to learn the discrepancy between the weak supervision and the ground truth in the target domain. Therefore, we can only use target data with accurate labels to estimate $\mathbf{d}$. Moreover, if we consider that the training loss $\widehat{L}(\mathbf{h})$ (which equals the average loss of all training samples) is hardly influenced by the sample quantity, and it is the same for the discrepancy between two domains~\cite{luo2020progressive}, we can split Eq.~(\ref{basic term of overall bound 1}) into two parts, where one part (denoted as $\Delta$) is not influenced by the sample quantity and the other part is related to the sample quantity. Based on Eq.~(\ref{pac_theorem_further}), these two parts can be written as follows:
\begin{equation}
\setlength{\abovedisplayskip}{3pt}
\setlength{\belowdisplayskip}{3pt}
\begin{aligned}
    R^t(\mathbf{h} \oplus \mathbf{d}) &\leq \Delta + 4\sqrt{\frac{K\!L_{\mathbf{d}}}{N_t}} + 4\sqrt{\frac{K\!L_{\mathbf{h}}}{N_s}} \\
    &\quad+ 12\sqrt{\frac{\ln{\frac{2N_t}{\delta}}}{N_t}} + 8\sqrt{\frac{\ln{\frac{2N_s}{\delta}}}{N_s}} \\ 
    \Delta &= 2\widehat{L}_t(\mathbf{h}^w) + \widehat{L}_t(\mathbf{d}) + \widehat{L}_s(\mathbf{h}) \\
    &\quad+ \widehat{L}_s(\mathbf{h}^w) + \mathcal{DD}(\mathbb{Q}^t_X, \mathbb{Q}^s_X)
    \label{basic term of overall bound 2}
\end{aligned}
\end{equation}
Here $K\!L_{\mathbf{d}}$ and $K\!L_{\mathbf{h}}$ denote KL divergences between trained $\mathbf{d}$, $\mathbf{h}$ and $\mathcal{H}$ respectively. According to Theorem 1, this KL divergence term is influenced by the training, especially impacted by the sample quantity. This is further discussed in Section 4 of the main paper.


\section{Dataset}
\label{Appendix-dataset}
The experiments are conducted on three application scenarios: the digits recognition with domain discrepancy (SVHN\cite{netzer2011reading}, MNIST\cite{deng2012mnist} and USPS\cite{hull1994database} digit datasets), object detection with domain discrepancy (VisDA-C\cite{peng2017visda}), and object detection without domain discrepancy (CIFAR-10\cite{krizhevsky2009learning}). 
\begin{myitemize}
    
    \item SVHN, MNIST, USPS: 
    SVHN dataset is a real-world image dataset that has 73,257 training examples obtained from house numbers in Google Street View images. MNIST is a widely used dataset for handwritten digit recognition and contains 60,000 training examples.  USPS database contains 7291 training images scanned from mail in a working post office. Every set is composed of 10 classes corresponding to 10 digit numbers. In the experiments, we randomly and non-repetitively sample 15,000 samples (each class includes 1,000 samples) from one domain as the source domain data, 1,000 samples from another domain as the target domain data, and another 2,000 samples as the validation data. We tested some combinations of adaptations chosen from these three datasets. All the images are resized to 32 $\times$ 32.
    
    
    \item VisDA-C:
    VisDA-C is a challenging object recognition dataset that contains 152,397 synthetic images and 55,388 real-world object images on 12 object classes. In the experiments, we randomly and non-repetitively sample 20,000 samples from synthetic images as the source domain data, 1,000 samples from real-world object images as the target domain data, and another 2,000 samples from real-world object images as the validation data. All the images are resized to 224 $\times$ 224.
    
    \item CIFAR-10
    CIFAR-10 is a popular object recognition benchmark with 50,000 training samples. We use this dataset here for testing our method under the scenario without domain discrepancy. Additionally, we randomly select 10,000 data samples from the dataset as the source domain data and another 1,000 data samples as the target domain data. All the images are resized to 32 $\times$ 32.

\end{myitemize}

\section{Additional Ablation Study}
\label{Appendix-ab}

\begin{table*}  
\small
\centering
\begin{tabular*}{16.6cm}{cccccccccccccc}  
\hline  
Method & knife & plane & bcycl & person & mcycl & car & truck & plant & bus & sktbrd & horse & train & Acc(\%)\\
\hline
$B_{wa}$ & 01.55 & 0.50 & 0.62 & 05.49 & 80.24 & 46.63 & 06.08 & 51.24 & 05.04 & 42.94 & 13.20 & 07.17 & 27.02\\
$B_{t}$ & 15.85 & 33.33 & 32.54 & 34.69 & 55.20 & 42.08 & 32.11 & 25.68 & 15.48 & 13.95 & 27.16 & 29.49 & 32.86\\
$B_{f_1}$ & 12.34 & 10.58 & 15.74 & 07.53 & 44.34 & 51.91 & 10.00 & 29.53 & 28.99 & 37.69 & 12.19 & 19.23 & 27.67\\
$B_{f_2}$ & 30.86 & 09.41 & 11.90 & 23.97 & 53.84 & 47.68 & 25.13 & 43.04 & 46.10 & 36.92 & 22.08 & 28.20 & 35.03\\
S+T & 00.05 & 03.84 & 03.34 & 00.00 & 14.24 & 86.16 & 07.16 & 14.42 & 16.78 & 02.67 & 00.00 & 00.00 & 21.56 \\
ENT & 22.07 & 33.85 & 42.79 & 10.18 & 34.84 & 45.07 & 37.61 & 45.19 & 20.58 & 10.08 & 13.99 & 27.53 & 31.51 \\
MME & 26.42 & 35.57 & 29.91 & 13.98 & 40.47 & 21.39 & 20.53 & 50.05 & 43.72 & 33.61 & 36.65 & 14.99 & 31.39 \\
FAN & 18.17 & 36.23 & 31.48 & 16.87 & 35.79 & 40.13 & 27.51 & 48.23 & 32.97 & 29.40 & 25.66 & 22.17 & 32.99 \\
Ours & 26.50 & 38.37 & 16.53 & 12.16 & 69.72 & 52.32 & 11.91 & 44.00 & 52.66 & 60.00 & 37.42 & 36.84 & 40.82 \\
\hline  
\end{tabular*}
\caption{The accuracy of different methods on the VisDA-C dataset with 12 classes. The number is measured in percentage. The accuracy of each class is at column 2 to column 13, and the overall accuracy is shown in the last column. This table shows results using a weak annotator that has worse performance than the model trained with target data only ($B_t$).
}

\label{result2-2}
\end{table*} 

In addition to the ablation experiments shown in our main paper, we also conduct additional studies that provide the experimental evidence on why we choose to estimate the distance of the optimal classifier for the target data and the  weak annotator, instead of learning the target task directly using the target data in Stage 2 of our algorithm (Algorithm 1 in the main paper) and then gives pseudo labels for the source data in Stage 3.

The experiment setting is the same as in the Section 5.5 of the main paper on the CIFAR-10 dataset. The algorithm for the ablation study is the same as Algorithm 1 in the main paper except for the following modifications: 1) The network $F_2$ will learn the classification result from the target data instead of learning the difference. In this way, $\Phi_2$ will only take the output feature from $\Phi_0$ as the input feature. 2) After finishing learning in Stage 2, when generating the new dataset in Stage 3, it becomes
\begin{equation}
\begin{aligned}
    D_{new} = \{(\mathbf{x}, y^{new}) | \mathbf{x} \in D_X, y^{new} = \Phi_0 \circ \Phi_2 (\mathbf{x})\} 
\end{aligned}
\setlength{\belowdisplayskip}{3pt}
\end{equation}
All other steps are the same as Algorithm 1. This additional ablation study actually follows the thought of assigning pseudo labels for all the data. Note that in the original paper of pseudo label~\cite{lee2013pseudo}, it is only for the unlabeled data. However the target data should also be re-labeled using the soft labels, following the ideas in~\cite{hinton2015distilling} -- this provides better performance and is the same approach as our method for this part.

The additional ablation study result is shown in Table~\ref{result-Ap}. The performance of the ablation method $B_{direct} =$ $57.70\%$ is lower than our result $61.71\%$, which indicates that learning the classification difference as in our Algorithm 1 is a better solution.

\begin{table*} 
\vspace{-0.1cm}
\small
\centering
\begin{tabular*}{14.4cm}{cccccccccccccc}  
\hline  
Method & plane & mobile & bird & cat & deer & dog & frog & horse & ship & truck & Acc(\%)\\
\hline
$B_{wa}$ & 43.18 & 65.68 & 28.13 & 25.93 & 29.00 & \textbf{\color{brown}46.15} & \textbf{\color{brown}83.91} & 41.76 & \textbf{72.12} & 51.06 & 48.96\\
$B_{direct}$ & 62.06 & 65.16 & \textbf{\color{brown}45.64} & 48.42 & 46.63 & 43.39 & 64.48 & 61.79 & 67.12 & 73.63 & 57.70\\
Ours & \textbf{\color{brown}65.52} & \textbf{\color{brown}82.61} & 39.79 & \textbf{\color{brown}48.45} & \textbf{\color{brown}57.36} & 43.60 & 67.39 & \textbf{\color{brown}65.32} & 70.42 & \textbf{\color{brown}78.89} & \textbf{\color{brown}61.71}\\
\hline  
\end{tabular*}
\caption{Additional experiment on whether we should choose to learn the classification difference or the target task directly  in Stage 2.
}
\label{result-Ap}
\end{table*} 

\begin{table*}  
\small
\centering
\begin{tabular*}{16.6cm}{cccccccccccccc}  
\hline  
Method & knife & plane & bcycl & person & mcycl & car & truck & plant & bus & sktbrd & horse & train & Acc(\%)\\
\hline
$B_{wa}$ & 41.89 & 53.03 & 34.17 & 46.67 & 72.59 & 55.42 & 07.07 & 47.13 & 24.84 & 40.00 & 42.85 & 25.32 & 42.14\\
$B_{t}$ & 15.85 & 33.33 & 32.54 & 34.69 & 55.20 & 42.08 & 32.11 & 25.68 & 15.48 & 13.95 & 27.16 & 29.49 & 32.86\\
$B_{f_1}$ & 31.71 & 40.31 & 11.20 & 22.30 & 64.09 & 39.45 & 10.00 & 44.74 & 39.05 & 18.82 & 31.10 & 23.23 & 33.56\\
$B_{f_2}$ & 34.14 & 08.13 & 24.19 & 26.71 & 63.30 & 55.73 & 25.13 & 57.89 & 37.86 & 44.69 & 50.60 & 40.25 & 42.84\\
S+T & 22.62 & 37.24 & 17.07 & 03.40 & 10.29 & 06.65 & 29.05 & 35.72 & 09.20 & 23.81 & 40.85 & 03.52 & 23.36 \\
ENT & 40.53 & 36.06 & 46.62 & 11.80 & 19.85 & 22.87 & 37.71 & 45.08 & 14.19 & 28.97 & 20.52 & 13.47 & 28.98 \\
MME & 21.60 & 44.95 & 17.05 & 24.88 & 57.13 & 36.62 & 16.72 & 18.65 & 36.57 & 28.93 & 20.83 & 13.11 & 29.31 \\
FAN & 32.71 & 50.13 & 14.72 & 20.09 & 60.78 & 53.44 & 20.60 & 42.34 & 20.07 & 17.65 & 23.91 & 33.25 & 34.47 \\
Ours & 43.37 & 59.54 & 13.49 & 21.76 & 67.43 & 60.92 & 21.28 & 51.97 & 58.08 & 18.60 & 40.24 & 40.38 & 45.06 \\
\hline  
\end{tabular*}
\caption{The accuracy of different methods on the VisDA-C dataset with 12 classes. The number is measured in percentage. The accuracy of each class is at column 2 to column 13, and the overall accuracy is shown in the last column. This table shows results using a weak annotator that has better performance than the model trained with target data only ($B_t$).
}

\label{result2-1}
\end{table*} 

\section{Network Architecture}
    \label{Appendix-Network-arch}
    
    In the digit experiments,
    the weak annotator is chosen as ResNet-18~\cite{he2016deep}. It is trained with randomly selected 10,000 data samples from the source dataset and 100 from the target dataset for 4 epochs.
    $\Phi_0$ is made from the VGG-19~\cite{simonyan2014very} network.
    $\Phi_1$ consists of three fully-connected layers, and the neuron number is set to [128, 64, 10].
    $\Phi_2$ consists of two fully-connected layers, and the neuron number is set to [512, 10].
    
    In the VisDA-C experiments, 
    The weak annotator is chosen as ResNet-50~\cite{he2016deep}. It is trained with randomly selected 10,000 data samples from both datasets for 10 epochs.
    $\Phi_0$ is made from the ResNet-50~\cite{simonyan2014very} network.
    $\Phi_1$ consists of three fully-connected layers, and the neuron number is set to [128, 64, 12].
    $\Phi_2$ consists of two fully-connected layers, and the neuron number is set to [1024, 10].
    
    In the CIFAR-10 experiments,
    The weak annotator is chosen as VGG-19~\cite{simonyan2014very}. It is trained with randomly selected 10,000 data samples from the dataset for 7 epochs.
    $\Phi_0$ is made from the VGG-19~\cite{simonyan2014very} network.
    $\Phi_1$ consists of three fully-connected layers, and the neuron number is set to [128, 64, 10].
    $\Phi_2$ consists of two fully-connected layers, and the neuron number is set to [64, 10].
    
\section{Training Settings}
    \label{Appendix-train}
    Generally, the scaling factor $\alpha$ in our algorithm is set to $1e-4$.
    The learning rate is selected from [1e-1, 1e-2, 1e-3, 1e-4, 1e-5], for the value with the best performance in experiments. The training epochs are empirically set as multiples of 10 and are selected for each experiment. We pre-run each experiment to determine the epoch value and stop training when the performance does not increase in the next 20 epochs to prevent over-fitting. 
    
    In the digit experiments, the training epochs in each training step is chosen as: ${ep}_1 = 90$, ${ep}_2 = 90$, ${ep}_3 = 40$, ${ep}_4 = 180$. The learning rate for experiment M $\to$ S is set to $1e-4$ and for other experiments set to $1e-5$. Training batch size is set to $128$. For the baseline $B_t$, it is trained for $90$ epochs, and the learning rate is $1e-5$ ($1e-4$ for M $\to$ S). For $B_{f_1}$, it is trained on the source data with weak labels for $90$ epochs and on the target data for $90$ epochs, and the learning rate is $1e-5$ ($1e-4$ for M $\to$ S). For $B_{f_2}$, it is trained on the source data with weak labels for $90$ epochs and on the target data for $90$ epochs, and the learning rate is $1e-5$ ($1e-4$ for M $\to$ S). Moreover, image augmentation techniques (provided by Torchvision.Transform) are applied for baselines $B_{t}$, $B_{f_1}$, $B_{t_2}$, and our approach. Other baselines use their original augmentation setting. We use the function in the Pytorch vision package for the implementation,
    and the images may be rotated from $-3$ to $3$ degree, or changed to gray-scale with a probability of 0.1.
    
    In the VisDA-C experiments, the training epochs in each training step is chosen as: ${ep}_1 = 90$, ${ep}_2 = 90$, ${ep}_3 = 40$, ${ep}_4 = 180$. The learning rate for experiment is set to $1e-5$. Training batch size is set to $128$. For the baseline $B_t$, it is trained for $90$ epochs, and learning rate is $1e-5$. For $B_{f_1}$, it is trained on the source data with weak labels for $90$ epochs and on the target data for $90$ epochs, and the learning rate is $1e-5$. For $B_{f_2}$, it is trained on the source data with weak labels for $90$ epochs and on the target data for $90$ epochs, and the learning rate is $1e-5$. The image augmentation techniques are also applied for baselines $B_{t}$, $B_{f_1}$, $B_{t_2}$, and our approach. Other baselines use their original augmentation setting. We similarly use the function in the Pytorch vision package for the implementation,
    and the images may be rotated from $-3$ to $3$ degree, or changed to gray-scale with a probability of 0.1, or horizontally flipped with a probability of 0.5.
    
    In the CIFAR-10 experiments, the training epochs in each training step is chosen as: ${ep}_1 = 40$, ${ep}_2 = 30$, ${ep}_3 = 70$, ${ep}_4 = 70$. The learning rate is set to $1e-3$. Training batch size is set to $128$. For the baseline $B_t$, it is trained for $70$ epochs, and the learning rate is $1e-3$. For $B_{f_1}$, it is trained on the source data with weak labels for $30$ epochs and on the target data for $40$ epochs, and the learning rate is $1e-3$. For $B_{f_2}$, it is trained on the source data with weak labels for $30$ epochs and on the target data for $40$ epochs, and the learning rate is $1e-3$. The image augmentation techniques are still applied to baselines $B_{t}$, $B_{f_1}$, $B_{t_2}$, and our approach. We use the function in the Pytorch vision package for the implementation,
    and the images are horizontally flipped with a probability of 0.5.

\section{Additional Experiments Details}
We provide more details about the accuracy of different methods on VisDA-C dataset, which is shown in Table~\ref{result2-1} and Table~\ref{result2-2}. In Table~\ref{result2-1}, we utilize a weak annotator that has better performance than the model trained with target data along $B_t$. And in Table~\ref{result2-2}, we employ a weak annotator that has worse performance than $B_t$. Both of them show that our method can provide a better performance boost compared to all baselines above.

\section{The impact of domain discrepancy}
\begin{figure}[htbp]
    \vspace{-0.3cm}
    \centering
    \includegraphics[width=0.5\textwidth]{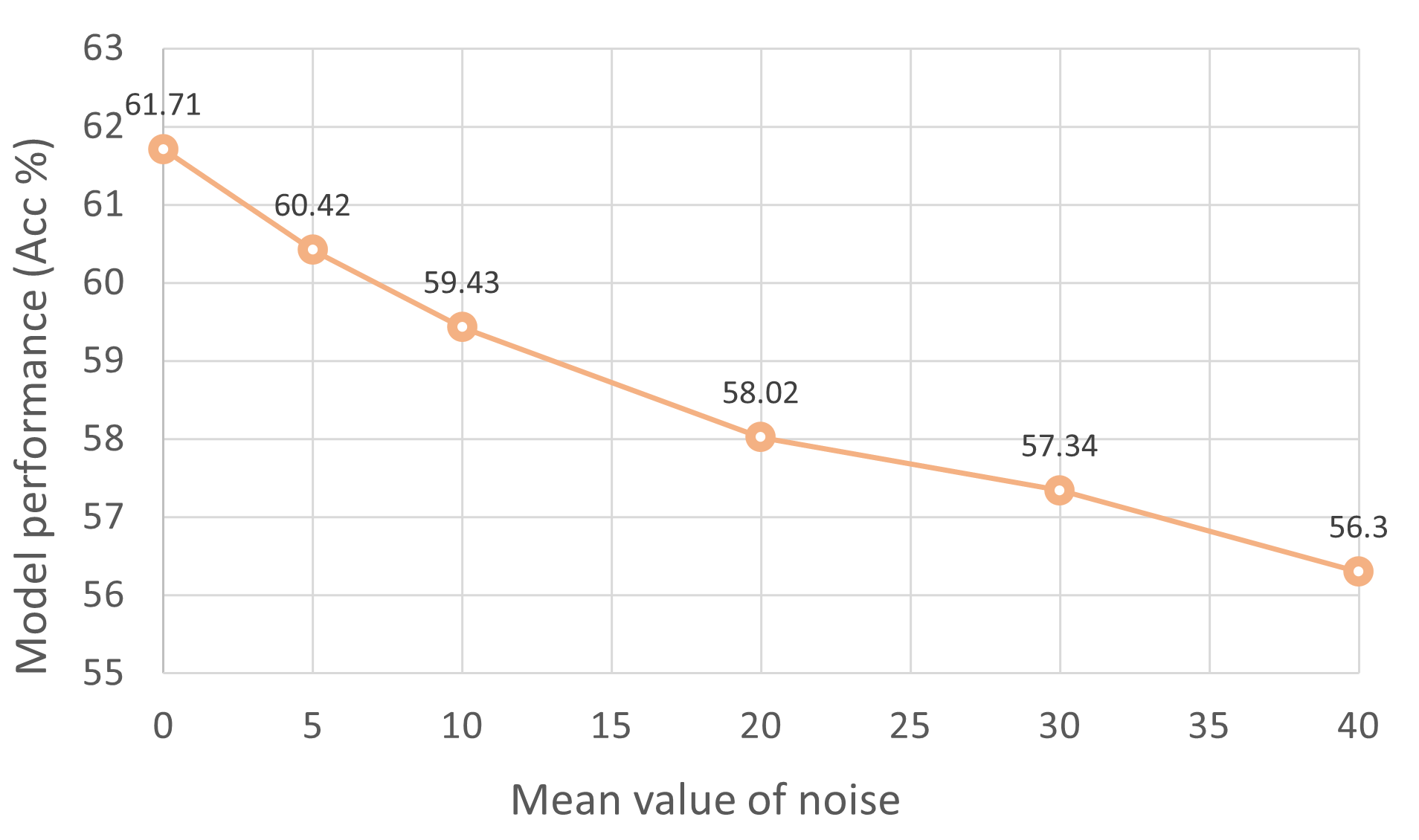}
    \vspace{-0.1cm}
    \caption{\small We add gaussian noise ($\sigma$=5.0) to CIFAR-10 on source data to create various data domains. The mean of noise reflects the domain discrepancy of source data and non-noisy target data.}
    \vspace{-0.3cm}
    \label{figure1}
\end{figure}

We also evaluate how the domain discrepancy will affect the model performance, and we conduct experiments on domain data with various levels of domain discrepancy. To be specific, we utilize gaussian noise ($\sigma$=5.0) with different level of mean value, and add it to the source data to create various data domains as the source domain. And the final model performance is shown in Figure~\ref{figure1}.

%% file: main.bbl
\begin{thebibliography}{10}\itemsep=-1pt

\bibitem{amit2018meta}
Ron Amit and Ron Meir.
\newblock Meta-learning by adjusting priors based on extended pac-bayes theory.
\newblock In {\em International Conference on Machine Learning}, pages
  205--214. PMLR, 2018.

\bibitem{belharbi2020deep}
Soufiane Belharbi, Ismail Ben~Ayed, Luke McCaffrey, and Eric Granger.
\newblock Deep active learning for joint classification \& segmentation with
  weak annotator.
\newblock In {\em Proceedings of the IEEE/CVF Winter Conference on Applications
  of Computer Vision}, pages 3338--3347, 2020.

\bibitem{cao2019theoretical}
Tianshi Cao, Marc Law, and Sanja Fidler.
\newblock A theoretical analysis of the number of shots in few-shot learning.
\newblock {\em arXiv preprint arXiv:1909.11722}, 2019.

\bibitem{chen2019closer}
Wei-Yu Chen, Yen-Cheng Liu, Zsolt Kira, Yu-Chiang~Frank Wang, and Jia-Bin
  Huang.
\newblock A closer look at few-shot classification.
\newblock In {\em International Conference on Learning Representations}, 2019.

\bibitem{chen2020self}
Xuxi Chen, Wuyang Chen, Tianlong Chen, Ye Yuan, Chen Gong, Kewei Chen, and
  Zhangyang Wang.
\newblock Self-pu: Self boosted and calibrated positive-unlabeled training.
\newblock In {\em International Conference on Machine Learning}, pages
  1510--1519. PMLR, 2020.

\bibitem{deng2012mnist}
Li Deng.
\newblock The mnist database of handwritten digit images for machine learning
  research [best of the web].
\newblock {\em IEEE Signal Processing Magazine}, 29(6):141--142, 2012.

\bibitem{dong2019semantic}
Jiahua Dong, Yang Cong, Gan Sun, and Dongdong Hou.
\newblock Semantic-transferable weakly-supervised endoscopic lesions
  segmentation.
\newblock In {\em Proceedings of the IEEE/CVF International Conference on
  Computer Vision}, pages 10712--10721, 2019.

\bibitem{dong2020can}
Jiahua Dong, Yang Cong, Gan Sun, Bineng Zhong, and Xiaowei Xu.
\newblock What can be transferred: Unsupervised domain adaptation for
  endoscopic lesions segmentation.
\newblock In {\em Proceedings of the IEEE/CVF conference on computer vision and
  pattern recognition}, pages 4023--4032, 2020.

\bibitem{dong2018tri}
WeiWang Dong-DongChen and Zhi-HuaZhou WeiGao.
\newblock Tri-net for semi-supervised deep learning.
\newblock In {\em International Joint Conferences on Artificial Intelligence},
  2018.

\bibitem{germain2016pac}
Pascal Germain, Francis Bach, Alexandre Lacoste, and Simon Lacoste-Julien.
\newblock Pac-bayesian theory meets bayesian inference.
\newblock In {\em Proceedings of the 30th International Conference on Neural
  Information Processing Systems}, pages 1884--1892, 2016.

\bibitem{ghosh2015making}
Aritra Ghosh, Naresh Manwani, and PS Sastry.
\newblock Making risk minimization tolerant to label noise.
\newblock {\em Neurocomputing}, 160:93--107, 2015.

\bibitem{goodfellow2016deep}
Ian Goodfellow, Yoshua Bengio, Aaron Courville, and Yoshua Bengio.
\newblock {\em Deep learning}, volume~1.
\newblock MIT press Cambridge, 2016.

\bibitem{grandvalet2004semi}
Yves Grandvalet and Yoshua Bengio.
\newblock Semi-supervised learning by entropy minimization.
\newblock In {\em Proceedings of the 17th International Conference on Neural
  Information Processing Systems}, pages 529--536, 2004.

\bibitem{gunst1977biased}
Richard~F Gunst and Robert~L Mason.
\newblock Biased estimation in regression: an evaluation using mean squared
  error.
\newblock {\em Journal of the American Statistical Association},
  72(359):616--628, 1977.

\bibitem{he2016deep}
Kaiming He, Xiangyu Zhang, Shaoqing Ren, and Jian Sun.
\newblock Deep residual learning for image recognition.
\newblock In {\em Proceedings of the IEEE conference on computer vision and
  pattern recognition}, pages 770--778, 2016.

\bibitem{hinton2015distilling}
Geoffrey Hinton, Oriol Vinyals, and Jeff Dean.
\newblock Distilling the knowledge in a neural network.
\newblock {\em arXiv preprint arXiv:1503.02531}, 2015.

\bibitem{huang2018auggan}
Sheng-Wei Huang, Che-Tsung Lin, Shu-Ping Chen, Yen-Yi Wu, Po-Hao Hsu, and
  Shang-Hong Lai.
\newblock Auggan: Cross domain adaptation with gan-based data augmentation.
\newblock In {\em Proceedings of the European Conference on Computer Vision
  (ECCV)}, pages 718--731, 2018.

\bibitem{hull1994database}
Jonathan~J. Hull.
\newblock A database for handwritten text recognition research.
\newblock {\em IEEE Transactions on pattern analysis and machine intelligence},
  16(5):550--554, 1994.

\bibitem{johnson2019survey}
Justin~M Johnson and Taghi~M Khoshgoftaar.
\newblock Survey on deep learning with class imbalance.
\newblock {\em Journal of Big Data}, 6(1):1--54, 2019.

\bibitem{kassam1978quantization}
S Kassam.
\newblock Quantization based on the mean-absolute-error criterion.
\newblock {\em IEEE Transactions on Communications}, 26(2):267--270, 1978.

\bibitem{kim2020attract}
Taekyung Kim and Changick Kim.
\newblock Attract, perturb, and explore: Learning a feature alignment network
  for semi-supervised domain adaptation.
\newblock In {\em European Conference on Computer Vision}, pages 591--607.
  Springer, 2020.

\bibitem{kingma2014adam}
Diederik~P Kingma and Jimmy Ba.
\newblock Adam: A method for stochastic optimization.
\newblock {\em arXiv preprint arXiv:1412.6980}, 2014.

\bibitem{kiryo2017positive}
Ryuichi Kiryo, Gang Niu, Marthinus C~du Plessis, and Masashi Sugiyama.
\newblock Positive-unlabeled learning with non-negative risk estimator.
\newblock {\em arXiv preprint arXiv:1703.00593}, 2017.

\bibitem{koenker2004quantile}
Roger Koenker.
\newblock Quantile regression for longitudinal data.
\newblock {\em Journal of Multivariate Analysis}, 91(1):74--89, 2004.

\bibitem{krizhevsky2009learning}
Alex Krizhevsky, Geoffrey Hinton, et~al.
\newblock Learning multiple layers of features from tiny images.
\newblock 2009.

\bibitem{lee2013pseudo}
Dong-Hyun Lee et~al.
\newblock Pseudo-label: The simple and efficient semi-supervised learning
  method for deep neural networks.
\newblock In {\em Workshop on challenges in representation learning, ICML},
  volume~3, 2013.

\bibitem{li2019dividemix}
Junnan Li, Richard Socher, and Steven~CH Hoi.
\newblock Dividemix: Learning with noisy labels as semi-supervised learning.
\newblock In {\em International Conference on Learning Representations}, 2019.

\bibitem{li2020model}
Rui Li, Qianfen Jiao, Wenming Cao, Hau-San Wong, and Si Wu.
\newblock Model adaptation: Unsupervised domain adaptation without source data.
\newblock In {\em Proceedings of the IEEE/CVF Conference on Computer Vision and
  Pattern Recognition}, pages 9641--9650, 2020.

\bibitem{liu2015classification}
Tongliang Liu and Dacheng Tao.
\newblock Classification with noisy labels by importance reweighting.
\newblock {\em IEEE Transactions on pattern analysis and machine intelligence},
  38(3):447--461, 2015.

\bibitem{loghmani2020positive}
Mohammad~Reza Loghmani, Markus Vincze, and Tatiana Tommasi.
\newblock Positive-unlabeled learning for open set domain adaptation.
\newblock {\em Pattern Recognition Letters}, 136:198--204, 2020.

\bibitem{long2016unsupervised}
Mingsheng Long, Han Zhu, Jianmin Wang, and Michael~I Jordan.
\newblock Unsupervised domain adaptation with residual transfer networks.
\newblock In {\em Advances in neural information processing systems}, pages
  136--144, 2016.

\bibitem{luo2020progressive}
Yadan Luo, Zijian Wang, Zi Huang, and Mahsa Baktashmotlagh.
\newblock Progressive graph learning for open-set domain adaptation.
\newblock In {\em International Conference on Machine Learning}, pages
  6468--6478. PMLR, 2020.

\bibitem{mansour2009domain}
Yishay Mansour, Mehryar Mohri, and Afshin Rostamizadeh.
\newblock Domain adaptation: Learning bounds and algorithms.
\newblock {\em arXiv preprint arXiv:0902.3430}, 2009.

\bibitem{mathieu2016disentangling}
Michael~F Mathieu, Junbo~Jake Zhao, Junbo Zhao, Aditya Ramesh, Pablo
  Sprechmann, and Yann LeCun.
\newblock Disentangling factors of variation in deep representation using
  adversarial training.
\newblock {\em Advances in neural information processing systems},
  29:5040--5048, 2016.

\bibitem{mcallester2003simplified}
David McAllester.
\newblock Simplified pac-bayesian margin bounds.
\newblock In {\em Learning theory and Kernel machines}, pages 203--215.
  Springer, 2003.

\bibitem{mcallester1999some}
David~A McAllester.
\newblock Some pac-bayesian theorems.
\newblock {\em Machine Learning}, 37(3):355--363, 1999.

\bibitem{natarajan2013learning}
Nagarajan Natarajan, Inderjit~S Dhillon, Pradeep Ravikumar, and Ambuj Tewari.
\newblock Learning with noisy labels.
\newblock In {\em NIPS}, volume~26, pages 1196--1204, 2013.

\bibitem{netzer2011reading}
Yuval Netzer, Tao Wang, Adam Coates, Alessandro Bissacco, Bo Wu, and Andrew~Y
  Ng.
\newblock Reading digits in natural images with unsupervised feature learning.
\newblock 2011.

\bibitem{neyshabur2017exploring}
Behnam Neyshabur, Srinadh Bhojanapalli, David McAllester, and Nati Srebro.
\newblock Exploring generalization in deep learning.
\newblock In {\em Advances in neural information processing systems}, pages
  5947--5956, 2017.

\bibitem{neyshabur2018pac}
Behnam Neyshabur, Srinadh Bhojanapalli, and Nathan Srebro.
\newblock A pac-bayesian approach to spectrally-normalized margin bounds for
  neural networks.
\newblock In {\em International Conference on Learning Representations}, 2018.

\bibitem{paul2020domain}
Sujoy Paul, Yi-Hsuan Tsai, Samuel Schulter, Amit~K Roy-Chowdhury, and Manmohan
  Chandraker.
\newblock Domain adaptive semantic segmentation using weak labels.
\newblock {\em arXiv preprint arXiv:2007.15176}, 2020.

\bibitem{peng2017visda}
Xingchao Peng, Ben Usman, Neela Kaushik, Judy Hoffman, Dequan Wang, and Kate
  Saenko.
\newblock Visda: The visual domain adaptation challenge.
\newblock {\em arXiv preprint arXiv:1710.06924}, 2017.

\bibitem{perez2018weakly}
F{\'a}bio Perez, R{\'e}mi Lebret, and Karl Aberer.
\newblock Weakly supervised active learning with cluster annotation.
\newblock {\em arXiv preprint arXiv:1812.11780}, 2018.

\bibitem{pouyanfar2018dynamic}
Samira Pouyanfar, Yudong Tao, Anup Mohan, Haiman Tian, Ahmed~S Kaseb, Kent
  Gauen, Ryan Dailey, Sarah Aghajanzadeh, Yung-Hsiang Lu, Shu-Ching Chen,
  et~al.
\newblock Dynamic sampling in convolutional neural networks for imbalanced data
  classification.
\newblock In {\em 2018 IEEE conference on multimedia information processing and
  retrieval (MIPR)}, pages 112--117. IEEE, 2018.

\bibitem{ranjan2017l2}
Rajeev Ranjan, Carlos~D Castillo, and Rama Chellappa.
\newblock L2-constrained softmax loss for discriminative face verification.
\newblock {\em arXiv preprint arXiv:1703.09507}, 2017.

\bibitem{rasmus2015semi}
Antti Rasmus, Harri Valpola, Mikko Honkala, Mathias Berglund, and Tapani Raiko.
\newblock Semi-supervised learning with ladder networks.
\newblock {\em arXiv preprint arXiv:1507.02672}, 2015.

\bibitem{roy2019unsupervised}
Subhankar Roy, Aliaksandr Siarohin, Enver Sangineto, Samuel~Rota Bulo, Nicu
  Sebe, and Elisa Ricci.
\newblock Unsupervised domain adaptation using feature-whitening and consensus
  loss.
\newblock In {\em Proceedings of the IEEE Conference on Computer Vision and
  Pattern Recognition}, pages 9471--9480, 2019.

\bibitem{saito2019semi}
Kuniaki Saito, Donghyun Kim, Stan Sclaroff, Trevor Darrell, and Kate Saenko.
\newblock Semi-supervised domain adaptation via minimax entropy.
\newblock In {\em Proceedings of the IEEE/CVF International Conference on
  Computer Vision}, pages 8050--8058, 2019.

\bibitem{saito2018maximum}
Kuniaki Saito, Kohei Watanabe, Yoshitaka Ushiku, and Tatsuya Harada.
\newblock Maximum classifier discrepancy for unsupervised domain adaptation.
\newblock In {\em Proceedings of the IEEE Conference on Computer Vision and
  Pattern Recognition}, pages 3723--3732, 2018.

\bibitem{shermin2020adversarial}
Tasfia Shermin, Guojun Lu, Shyh~Wei Teng, Manzur Murshed, and Ferdous Sohel.
\newblock Adversarial network with multiple classifiers for open set domain
  adaptation.
\newblock {\em IEEE Transactions on Multimedia}, 2020.

\bibitem{simonyan2014very}
Karen Simonyan and Andrew Zisserman.
\newblock Very deep convolutional networks for large-scale image recognition.
\newblock {\em arXiv preprint arXiv:1409.1556}, 2014.

\bibitem{snell2017prototypical}
Jake Snell, Kevin Swersky, and Richard Zemel.
\newblock Prototypical networks for few-shot learning.
\newblock In {\em Proceedings of the 31st International Conference on Neural
  Information Processing Systems}, pages 4080--4090, 2017.

\bibitem{tang2020discriminative}
Hui Tang and Kui Jia.
\newblock Discriminative adversarial domain adaptation.
\newblock In {\em AAAI}, pages 5940--5947, 2020.

\bibitem{triantafillou2019meta}
Eleni Triantafillou, Tyler Zhu, Vincent Dumoulin, Pascal Lamblin, Utku Evci,
  Kelvin Xu, Ross Goroshin, Carles Gelada, Kevin Swersky, Pierre-Antoine
  Manzagol, et~al.
\newblock Meta-dataset: A dataset of datasets for learning to learn from few
  examples.
\newblock In {\em International Conference on Learning Representations}, 2019.

\bibitem{tzeng2014deep}
Eric Tzeng, Judy Hoffman, Ning Zhang, Kate Saenko, and Trevor Darrell.
\newblock Deep domain confusion: Maximizing for domain invariance.
\newblock {\em arXiv preprint arXiv:1412.3474}, 2014.

\bibitem{van2007experimental}
Jason Van~Hulse, Taghi~M Khoshgoftaar, and Amri Napolitano.
\newblock Experimental perspectives on learning from imbalanced data.
\newblock In {\em Proceedings of the 24th international conference on Machine
  learning}, pages 935--942, 2007.

\bibitem{wang2021addressing}
Lixu Wang, Shichao Xu, Xiao Wang, and Qi Zhu.
\newblock Addressing class imbalance in federated learning.
\newblock In {\em Proceedings of the AAAI Conference on Artificial
  Intelligence}, volume~35, pages 10165--10173, 2021.

\bibitem{wang2020generalizing}
Yaqing Wang, Quanming Yao, James~T Kwok, and Lionel~M Ni.
\newblock Generalizing from a few examples: A survey on few-shot learning.
\newblock {\em ACM Computing Surveys (CSUR)}, 53(3):1--34, 2020.

\bibitem{wilson2020survey}
Garrett Wilson and Diane~J Cook.
\newblock A survey of unsupervised deep domain adaptation.
\newblock {\em ACM Transactions on Intelligent Systems and Technology (TIST)},
  11(5):1--46, 2020.

\bibitem{wilson2020multi}
Garrett Wilson, Janardhan~Rao Doppa, and Diane~J Cook.
\newblock Multi-source deep domain adaptation with weak supervision for
  time-series sensor data.
\newblock In {\em Proceedings of the 26th ACM SIGKDD International Conference
  on Knowledge Discovery \& Data Mining}, pages 1768--1778, 2020.

\bibitem{xu2021learningbased}
Shichao Xu, Yangyang Fu, Yixuan Wang, Zheng O'Neill, and Qi Zhu.
\newblock Learning-based framework for sensor fault-tolerant building hvac
  control with model-assisted learning, 2021.

\bibitem{xu2019positive}
Yixing Xu, Yunhe Wang, Hanting Chen, Kai Han, Chunjing Xu, Dacheng Tao, and
  Chang Xu.
\newblock Positive-unlabeled compression on the cloud.
\newblock {\em arXiv preprint arXiv:1909.09757}, 2019.

\bibitem{yi2017semismooth}
Congrui Yi and Jian Huang.
\newblock Semismooth newton coordinate descent algorithm for elastic-net
  penalized huber loss regression and quantile regression.
\newblock {\em Journal of Computational and Graphical Statistics},
  26(3):547--557, 2017.

\bibitem{zellinger2017central}
Werner Zellinger, Thomas Grubinger, Edwin Lughofer, Thomas Natschl{\"a}ger, and
  Susanne Saminger-Platz.
\newblock Central moment discrepancy (cmd) for domain-invariant representation
  learning.
\newblock {\em arXiv preprint arXiv:1702.08811}, 2017.

\bibitem{zhou2018brief}
Zhi-Hua Zhou.
\newblock A brief introduction to weakly supervised learning.
\newblock {\em National science review}, 5(1):44--53, 2018.

\end{thebibliography}
